\documentclass[letterpaper]{article} % DO NOT CHANGE THIS
\usepackage[preprint]{aaai2027}
\usepackage[hyphens]{url}  % DO NOT CHANGE THIS
\usepackage{graphicx} % DO NOT CHANGE THIS
\usepackage{natbib}  % DO NOT CHANGE THIS AND DO NOT ADD ANY OPTIONS TO IT
\usepackage{caption} % DO NOT CHANGE THIS AND DO NOT ADD ANY OPTIONS TO IT
\usepackage{algorithm}
\usepackage{algorithmic}

\usepackage{newfloat}
\usepackage{listings}
\DeclareCaptionStyle{ruled}{labelfont=normalfont,labelsep=colon,strut=off} % DO NOT CHANGE THIS
\floatstyle{ruled}
\newfloat{listing}{tb}{lst}{}
\floatname{listing}{Listing}

\usepackage{booktabs}

\usepackage{multirow}
\usepackage{amsmath}
\usepackage{booktabs} %for table midrule etc.
\usepackage{tcolorbox} %for tcolorrbox
\usepackage{algorithm}
\usepackage{algorithmic}
\usepackage{changepage} % for adjustwidth*
\usepackage{subcaption} %for subfigure
\usepackage[table]{xcolor} % Enables row coloring in tables
\usepackage{amsthm} %proof/theorem stuff
\usepackage{amssymb} %for mathbb
\usepackage{tabularx} %appn table etc
\usepackage{longtable}%^
\usepackage{makecell}
\usepackage{multibib}
\newcites{app}{Appendix References}
 \usepackage{hyperref}

\title{\href{https://veracitea.github.io/PAWS/}{\includegraphics[height=0.9em]{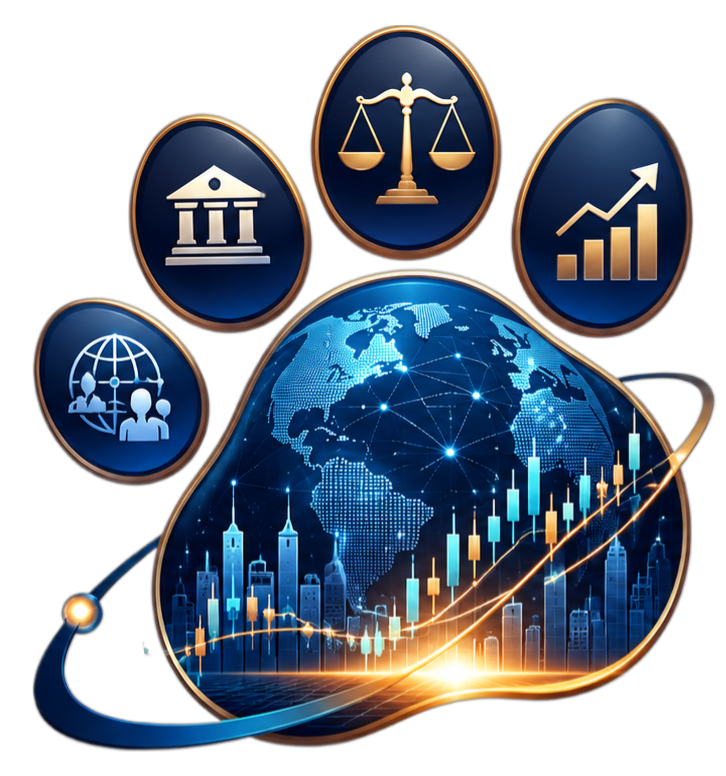}\includegraphics[height=0.8em]{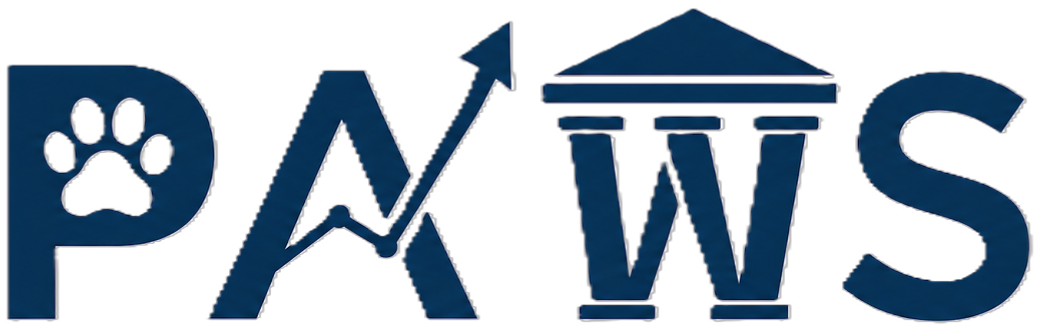}}: Policy-driven Agentic World Simulation}
\author{
    Tiviatis Sim,\textsuperscript{\rm 1}\corresponding, Woon Jia Hui\textsuperscript{\rm 1}, Xinming Gao\textsuperscript{\rm 2}, Chen Gao\textsuperscript{\rm 2}, Fengbin Zhu\textsuperscript{\rm 1}, Zheng Huanhuan\textsuperscript{\rm 3}\\
    Chua Tat Seng\textsuperscript{\rm 1}, Kenji Kawaguchi\textsuperscript{\rm 1}
 }
\affiliations{
    \textsuperscript{\rm 1}National University of Singapore,
    \textsuperscript{\rm 2} Tsinghua University,
    \textsuperscript{\rm 3} City University of Hong Kong\\
    tiviatis@u.nus.edu
}

\begin{document}

\maketitle

\begin{abstract}
Policy interventions propagate through public communication, institutional decisions, and stakeholder responses, yet datasets for financial multi-agent simulation rarely connect these processes to temporally aligned historical evidence. We introduce \href{https://veracitea.github.io/PAWS/}{\textbf{PAWS}, a \textbf{P}olicy-driven \textbf{A}gentic \textbf{W}orld \textbf{S}imulation dataset} covering 36 verified U.S. financial and economic policy episodes, 12,727 policy-linked news records, and 65,291 source-grounded stakeholder actions. Each action is linked to its supporting news and represented by a multi-layer event frame capturing its interaction mode, financial-action family and subtype, semantic attributes, and conditional mappings to external taxonomies. Entities are resolved to normalized organizations, and actions are aligned with daily market-return context to support policy-agent simulation replay. On 2,522 stratified action samples, independent AI and human reviewers achieved 89.4\% initial agreement on interaction mode, with disagreements subsequently adjudicated. Case studies of the 2008 short-selling ban and 2001 decimalization recover documented policy timelines and associated market patterns across both dense and sparse news settings. A replay study further shows that high accuracy can mask failure to detect rare stakeholder actions, identifying action timing and calibration as central challenges. PAWS provides an auditable substrate for evaluating agent influence, policy-response cascades, and action-outcome alignment in historically grounded financial simulations.
\end{abstract}


\section{Introduction}

Financial crises are driven not only by prices and balance sheets, but also by policy announcements, institutional responses, public communication, and rapidly changing expectations. Classical agent-based finance models can reproduce stylized market dynamics, but their agents operate over numerical states and hand-coded rules that ignore natural language data~\citep{Lebaron2006,FarmerFoley2009}; financial language models and trading-agent frameworks can process news and other textual signals, but are rarely evaluated in historical settings where policy intent, stakeholder actions, and market context are aligned over time~\citep{Liuetal2023FINGPT,liuetal2025quantagents,Sun2023TradeMaster,Liu_2021FinRL}. As a result, it remains difficult to ask whether simulated agents react plausibly to a real policy shock, whether their actions match the behaviour of relevant institutions, or whether a simulation captures the difference between an intended policy effect and the outcome later observed in the market.

We introduce \href{https://veracitea.github.io/PAWS/}{\textbf{PAWS}, a \textbf{P}olicy-driven \textbf{A}gentic \textbf{W}orld \textbf{S}imulation dataset} for studying policy-centered multi-agent behavior in financial markets. PAWS uses discrete U.S. financial and economic policy interventions whose intended goals and real market effects are grounded in finance research. For each policy, PAWS aligns policy metadata with dated news articles, extracted stakeholder actions, normalized action types, entity/organization mappings and daily market-return context. This allows researchers to move beyond isolated news classification/price prediction and instead evaluate agents in temporally grounded policy environments.

The current \href{https://veracitea.github.io/PAWS/}{PAWS database} contains 36 verified policy episodes, 301
policy-specific query keys, 12,727 policy-linked news rows, 65,291 extracted
action rows, and 65,291 corresponding action-frame rows. The core design choice in PAWS is to represent policy response as a structured
resource rather than as a single label. Each extracted action is grounded in
one or more news articles and is represented with a multi-layer event frame:
an interaction mode, a financial-action family, a family-constrained subtype,
modality and status attributes, and conditional crosswalks to external event
taxonomies.

The main contributions of this work can be summarized as follows.
% itemize
\begin{itemize}
	\item We introduce the first cohesive policy-driven multi-agent simulation (MAS) dataset that aligns historical policy shocks, policy-relevant news, stakeholder actions, and market-return context.
	\item We provide dataset analysis and a focused validation study on class labelling and action validity while presenting case studies to check if news events match academic conclusion of several financial events.
    \item We provide resources, with the necessary pipeline for replicating the dataset collection and related experiments, as well as an example on how PAWS can be used to do a simulation study.
    \item Unlike prior financial or political simulation, this dataset aligns verified financial-policy episodes with news evidence, stakeholder actions, and market data. Moreover, it is designed to allow both traditional MAS methods and modern LLMMAs.
\end{itemize}
%%%%%%%%%%%%%%%%%%%%%%%%%%%%%%%%

\begin{figure*}[t]
  \centering
  \includegraphics[width=1\textwidth, trim={0cm 0cm 0cm 0cm},clip]{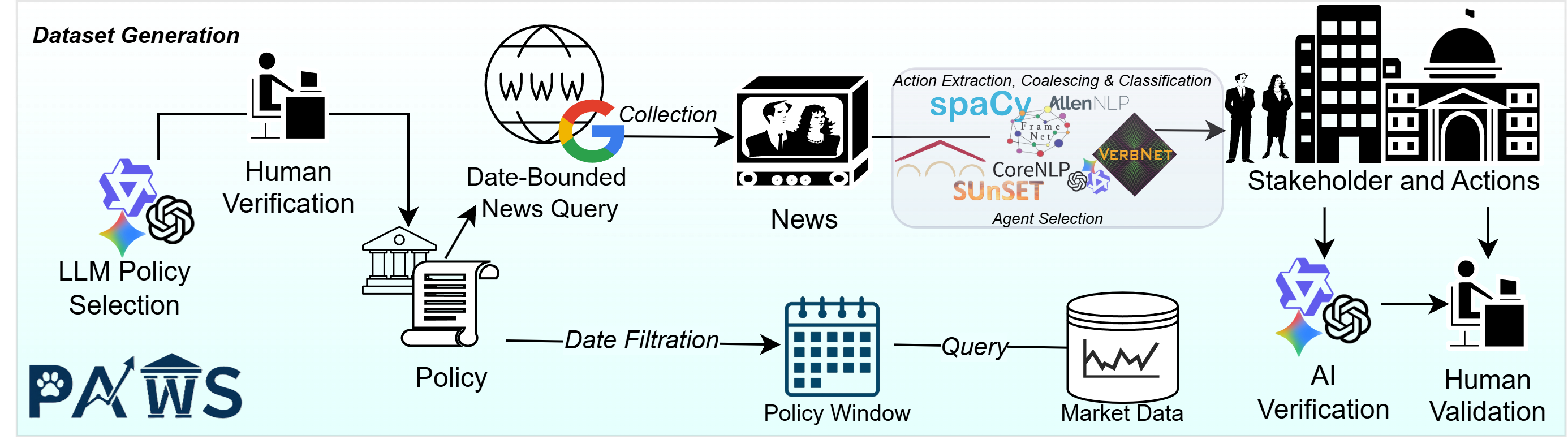}
  \caption{\small{ 
PAWS construction framework. Policy-centered news is collected over event windows, converted into stakeholder action records, resolved to normalized organizations, aligned with market context, and structured into replay-ready agent-day panels for MAS.
}}
\vspace{-3mm}
  \label{fig:framework}
\end{figure*}
%%%%%%%%%%%%%%%%%%%%%%%%%%%%%%%%

\section{Related Works} \label{ssec:related-work}

To contextualize PAWS, we draw upon three distinct but complementary lines of research. First, we examine financial and LLM-based agent simulations to highlight the growing need for grounded, historically accurate policy environments. Second, we review financial text and event resources to identify the gap in datasets that explicitly link policy interventions with structured stakeholder actions. Finally, we discuss literature on policy shocks and market responses, demonstrating how PAWS provides the structured data necessary to bridge qualitative policy analysis with quantitative market outcomes.

\paragraph{Financial and LLM-based agent simulation}
Traditional Agent-Based Computational Finance models are structurally robust for demonstrating systemic risk and policy failures during crises but are inherently text-blind~\citep{Lebaron2006,FarmerFoley2009}. Modern simulation platforms like ABIDES, FinRL, and TradeMaster enable high-fidelity market simulation and trading-agent research~\citep{Byrd2019ABIDES,Liu_2021FinRL,Sun2023TradeMaster}, and LLM-based agents such as FinGPT, QuantAgents, and TradingAgents have advanced reasoning and domain-specific collaborative trading~\citep{Park2023GenerativeAgents,Zhou2023SOTOPIA,Liuetal2023FINGPT,liuetal2025quantagents,Xiao2024TradingAgents}. However, these existing systems lack curated historical evidence linking policy announcements, policy intent, stakeholder behavior, and realized market context.

\paragraph{Financial text, event, and timeline resources}
Financial NLP datasets support tasks such as numerical reasoning, relation extraction, and event extraction over reports, news, and earnings calls~\citep{chen2022FinQA,Sharma2023FinRED,zheng-etal-2019-doc2edag,Huang2024FINEED}. These resources provide valuable supervised targets for extracting finance-relevant facts, but their units are typically documents, relations, or event mentions rather than policy episodes with explicit intervention windows. Broader event resources such as GDELT, ICEWS, and EventRegistry provide large-scale actor-event streams, while timeline summarization work studies how to identify and order salient events across document collections~\citep{WardEtAl2013GDELTICEWS,KwakAn2016GDELTEventRegistry,hu-etal-2024-moments,sunset}. While useful for extracting or organizing events, they are not curated around financial-policy interventions and do not jointly preserve policy motivation, stakeholder action traces and market outcomes; PAWS bridges policy intent, news evidence, stakeholder actions, and market context  as a single auditable resource.

\paragraph{Policy shocks and market responses}
Finance research commonly uses event studies to estimate the market impact of discrete economic events~\citep{MacKinlay1997EventStudies}. Work on economic policy uncertainty and monetary-policy event-study databases further demonstrate the importance of systematically connecting policy communication to financial-market responses~\citep{BakerBloomDavis2016EPU,AcostaEtAl2025USMPD}. PAWS complements this literature: rather than estimating a single treatment effect, it provides structured, temporally grounded evidence for studying how policy shocks propagate through stakeholders, actions, and downstream market context. This design enables PAWS to be used for studying how stakeholders respond to financial interventions, and a basis for downstream replay or simulation studies 

A more detailed comparison between PAWS and existing work is provided in Appendix~\ref{app:related-comparison}.

\section{Methodology}
\label{methodology}
PAWS is constructed as a policy-centered event resource for MAS. The pipeline selects policy shocks episodes, retrieves dated news within policy windows, extracts and coalesces stakeholder actions, maps each action into
a multi-layer event frame, resolves entities, and links the resulting records to daily financial market context.

\subsection{Dataset Generation}
\label{sec:dataset-construction}

\paragraph{Policy Selection}
While numerous financial policies are enacted annually, not all of them produce measurable or separable market effects. PAWS therefore focuses on policies that satisfy three criteria: the episode has an identifiable announcement or implementation date, it is relevant to financial markets or financial intermediation, and it has enough public documentation to support external verification. The current database snapshot contains 36 verified policy
episodes and 301 policy-specific query keys.

Candidate policies are initially compiled with \texttt{GPT-5.5}, prompted to act as a tenured finance professor and to prioritize measurable policy impacts. These candidates are then manually verified against reputable finance research and official policy documentation, then manually checked for dates, scope, and policy description. Each retained policy is recorded with its announcement window, policy source, intended effect, and documented actual outcome. The policies' intended outcomes, actual impacts, prompt template, and an alternative underperforming policy-identification variant are documented in Appendices~\ref{appn:prompts}, \ref{appn:PolicyInfo}, and \ref{appn:traderLLM}.

\paragraph{Market Data}
Market data is retrieved from the publicly available financial dataset 
factors, industry portfolios, and size portfolios \citep{fama1993commonfactors,fama1997industry,french_data}. The market tables are not used as labels for action extraction. Instead, they provide contemporaneous context for later analysis, including event-window summaries, descriptive comparisons across industries, and replay tasks that condition stakeholder behavior on daily
market state.

\paragraph{News Collection}
 For each verified policy, we define a policy window from the policy start date to its documented end date. When an end date is unavailable, we use a fixed post-announcement window and report this choice in the dataset analysis. We then issue date-bounded Google News queries using policy-specific keywords and source information stored in \texttt{PolicyKey}. Queries are run over daily intervals so that retrieved articles can be aligned with the policy timeline.

The retrieval stage resolves Google News links to canonical article URLs and stores the publication date, headline, article body, source domain, and retrieval count. If a canonical article is unavailable, we attempt to recover the page through the Internet Archive's Wayback Machine. Articles that cannot be fully recovered are retained only when their metadata is sufficient for provenance and are flagged for manual inspection and backfilling. The accepted records are stored in the \texttt{News} table with \texttt{policy\_id}, \texttt{date}, \texttt{headline}, \texttt{content}, \texttt{source}, and \texttt{count}. Duplicate source-date records are consolidated by incrementing \texttt{count}, with questionable or policy-irrelevant articles removed through human-in-the-loop review. This design preserves article provenance and allows every downstream action to be traced back to the specific policy-relevant sources from which it was extracted.

\paragraph{Action Extraction, Coalescing and Classification}
The action-extraction stage identifies stakeholders and the actions attributed to them in policy-window news. Each extracted row stores an actor string,  acting entity, normalized organization identifier when available, action text, date, source news identifiers, location, legacy action type and a structured event-frame representation. Some articles may be relevant to multiple policies, thus articles are grouped by canonical source before extraction. The \texttt{GPT-5.5} extraction model is then called once per source group and returns a structured JSON list of actions, which are linked back to all associated \texttt{news\_ids} and \texttt{policy\_ids}.

PAWS represents each extracted action as a structured event frame rather than a single flat class. For an action mention $a$, the event frame can be summarized as,

\begin{equation}
  \begin{split}
    \mathcal{F}(a)
    &=
    \langle
      m_a,\,
      f_a,\,
      s_a;\,
      x^{\mathrm{CAMEO}}_a,\,
      x^{\mathrm{Fed}}_a,\,
      x^{\mathrm{ACE}}_a,\,
      x^{\mathrm{Mon}}_a, \\
    &\quad
      x^{\mathrm{iMaPP}}_a,\,
      x^{\mathrm{IMF}}_a;\,
      \mathbf{z}_a
    \rangle ,
  \end{split}
  \label{eq:event-frame}
\end{equation}

where $m_a$ is the interaction mode, $f_a$ is the financial-action
family, $s_a$ is a family-constrained subtype, the six $x_a$ terms are
independently gated crosswalks, and $\mathbf{z}_a$ contains attributes such as modality, status, direction, actor role, target, instrument, amount, and evidence span, and $\mathbf{x}_a$ contains conditional crosswalks to external event or policy taxonomies.  

To generate the above, the \texttt{GPT-5.5}-generated label provides a context-sensitive interpretation of the full source group. Second, AllenNLP, OpenIE, and spaCy perform semantic role labelling (SRL) and are used to recover predicate-argument structures and classify action predicates~\citep{gardner2018allennlp,honnibal2020spacy}. Third, formal semantic resources including FrameNet, VerbNet, and FIBO provide ontology-grounded evidence for the same action~\citep{baker1998berkeley,schuler2005verbnet,edmcouncilfibo}. The full multi-layer event-frame categories can be accessed in Appendices~\ref{appn:frameclass} and \ref{app:framedes}.

Next, coreference resolution and organization identification are applied to extracted entities, since the same stakeholder may appear under different abbreviations or aliases. We build a knowledge graph from Wikidata~\citep{wikidata} to map these variants to normalized organizations and to capture multilingual aliases and transliterations. The specifics of the knowledge-graph construction are provided in the supplementary appendix.

Finally, raw extracted actions are coalesced into a daily action table. Records with the same normalized organization, event date, and action type are merged into a single \texttt{DailyAction} row. During coalescing, PAWS preserves the union of linked policy IDs, news IDs, and raw action IDs, and assigns sentiment by untied plurality when multiple source records are available. The resulting daily table is the primary temporal substrate used for agent selection and simulation.

\paragraph{Actor and Action Quality Control}
\label{agent_selection}
The actor classification is completed by a primary annotator, followed by verification from another annotator. All disagreements will be discussed and resolved. All stakeholder data undergoes idempotency check after classification.

However, due to the scale of the action set, action classification proceeds in two stages: AI verification and human validation. \texttt{Sol-5.6} was used as the AI verifier, where random class-stratified samples were selected; balanced combinations across differing classes were selected. The human verifier looks into the same samples, fills in the event-frame and cross-checks with the AI's output. For difficult samples with disagreements, the human will reassess the samples before attaching the golden label.

All samples and adjudication will be available in the resources.

\subsection{Derived Replay View}
We have provided a simple full-process replay for the September 2008 short-selling ban within this paper. In this experimental setup, each selected stakeholder is treated as an agent capable of individual decision making, and each policy-agent-day tuple is a decision point. The target label is one of \emph{no action}, \emph{Communication}, or \emph{Direct Intervention}. We utilized expressive replay models: a majority baseline, a per-agent Markov transition baseline~\citep{Rabiner1989HMM}, a network-threshold agent-based model inspired by threshold models of collective behavior~\citep{Granovetter1978Threshold}, a hybrid temporal-network agent-based model, a structured classifier, and a bounded \texttt{GPT-4o} LLM-agent replay. More information on the design can be found in Appendix~\ref{app:simulation_diagnostics}.

\section{Dataset Overview} 
\label{sec:dataset}
\begin{figure}[h]
    \centering
    \includegraphics[width=0.99\linewidth]{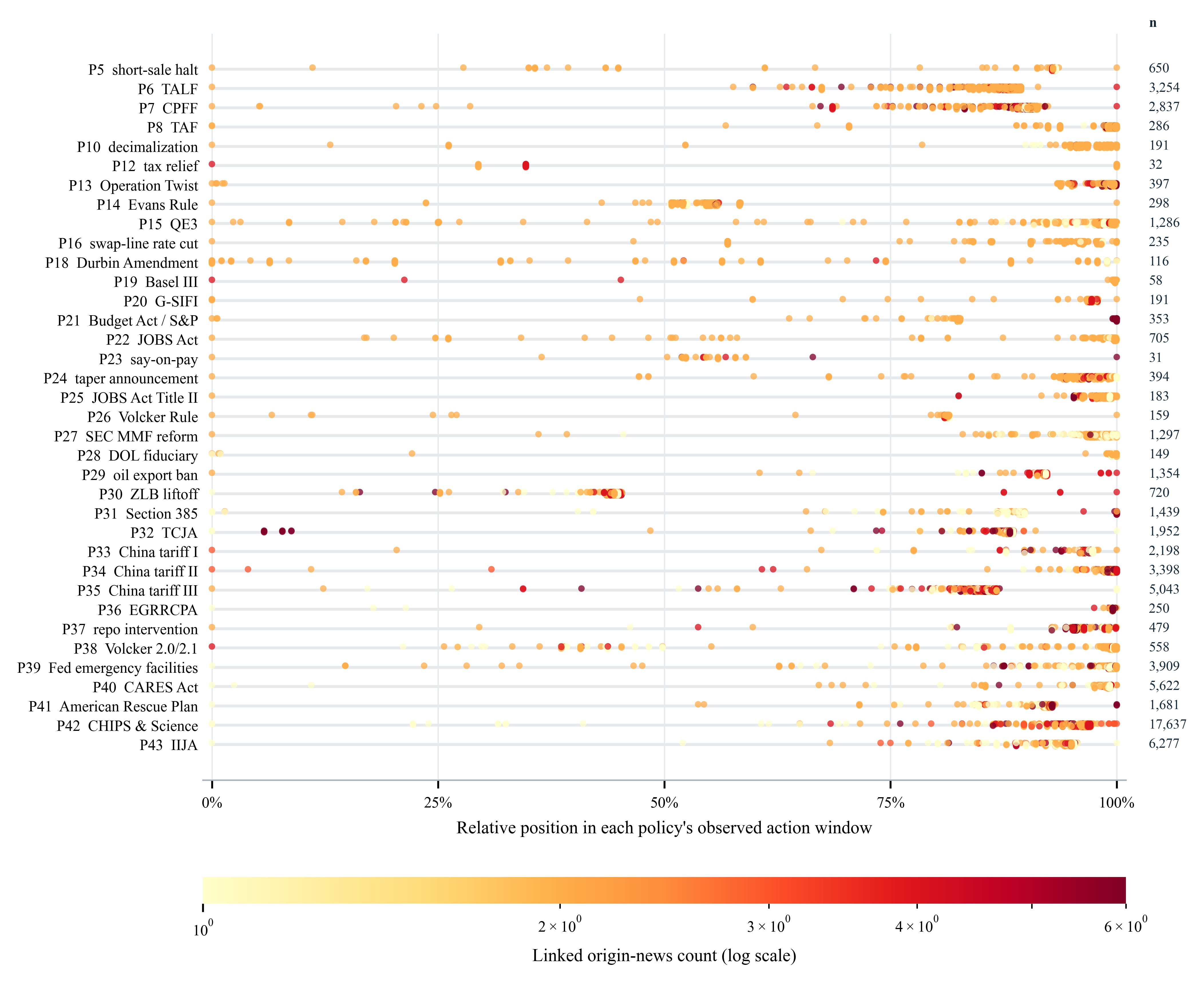}
    \caption{The proportion of actions done by different types of actors for each policy.}
    \label{fig:timeline}
\end{figure}

\begin{figure}[h]
    \centering
    \includegraphics[width=0.99\linewidth]{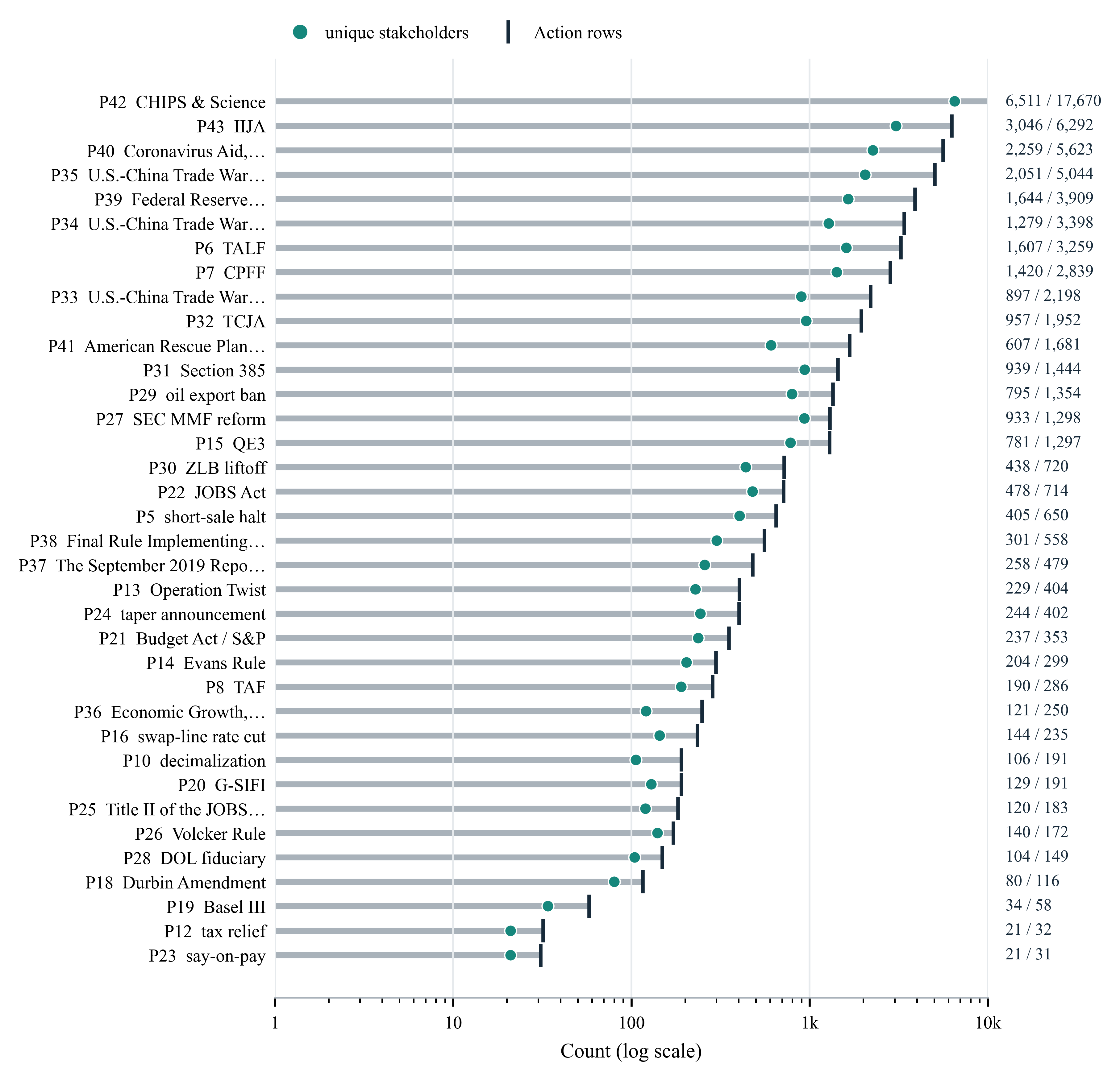}
\caption{Policy-level stakeholder diversity relative to corpus size. Circles denote unique normalized stakeholders, vertical ticks denote Action rows, and triangles denote ActionFrame rows when the counts differ. Counts are shown on a logarithmic scale.}
    \label{fig:person}
\end{figure}

PAWS contains 36 verified policy
episodes, 301 policy-specific query keys, 12,727 policy-linked news rows, 65,291 extracted action rows, and 65,291
corresponding action-frame rows. Figure~\ref{fig:timeline} showcases the timelines of all 36 policies, and Figure~\ref{fig:person} indicates the relative stakeholders to action distribution across policies. The most frequent stakeholders can be found in the appendix (Figure~\ref{fig:stakePop}).

\subsection{Multi-Layer Event-Frame Actions}
\label{sec:event-frame-representation}

We represent each extracted action as a structured event frame rather than as a single mutually exclusive label.  The design separates (i) the way in which an event is realized, (ii) its domain-level financial function, and (iii) its fine-grained mechanism.  Six external crosswalk systems then expose compatible views of the same event for political-event, monetary-policy, macroprudential-policy, and crisis-response analyses. The motivation of each event-frame can be found in Appendix~\ref{app:motivation}.

\begin{figure}[h]
    \centering
    \includegraphics[width=0.99\linewidth]{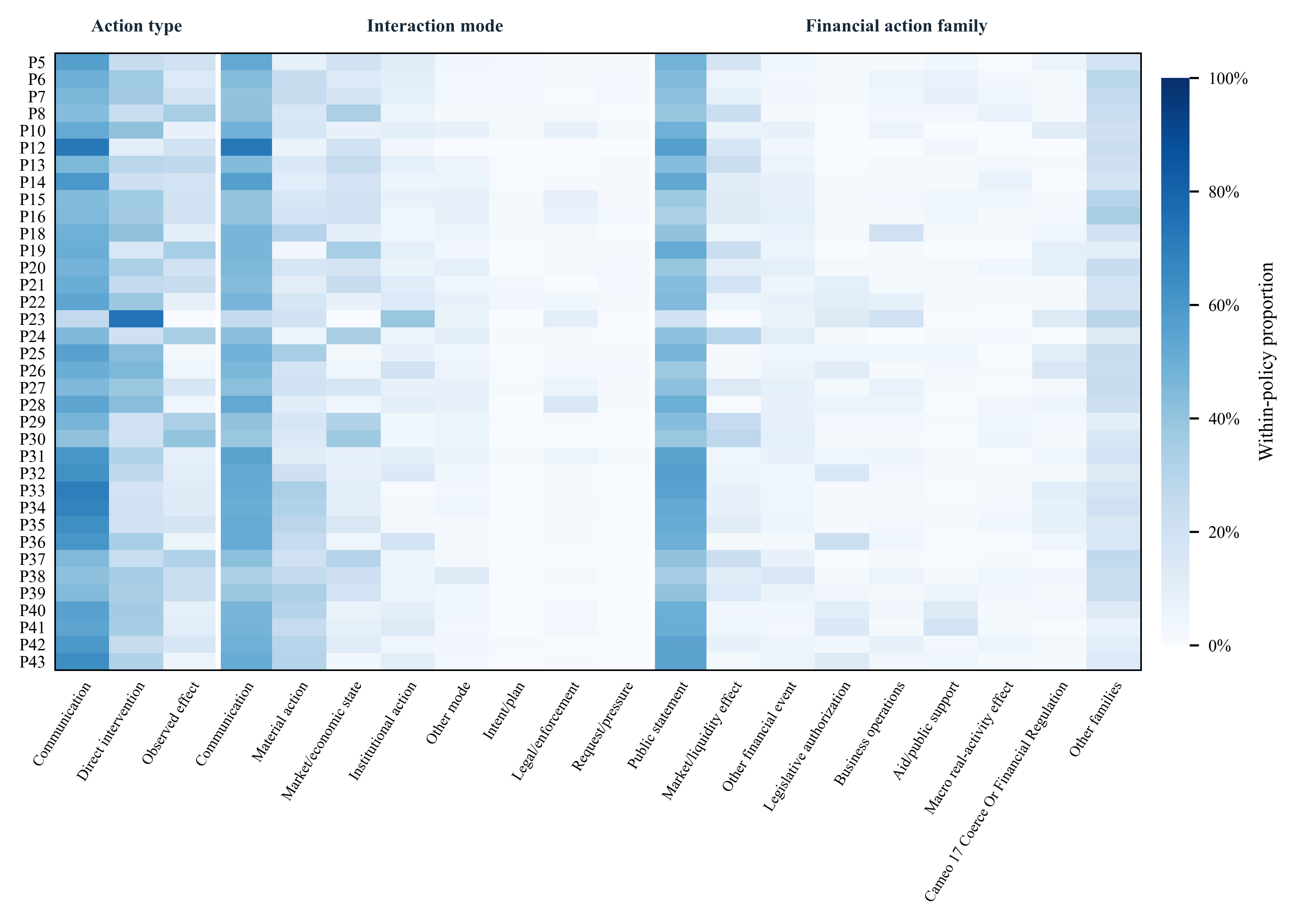}

   \caption{Within-policy composition of action representations across 36 policy episodes. Rows denote policies, while columns group the corresponding action type, interaction mode, and financial action family. Each cell shows the proportion of actions assigned to that class, normalized separately within each representation layer; darker blue indicates greater prevalence. The family panel presents the eight most frequent financial-action families, with all remaining categories pooled as \emph{Other families}.}
    \label{fig:class_distribution}
\end{figure}

\paragraph{Interaction Mode}
The 8-valued interaction mode records how the event is performed or where it lies in its realization lifecycle.  It distinguishes communicative reporting, an actor's plan or commitment, a formal institutional act, directed influence or pressure, a material resource transfer, use of a policy instrument, an organizational/business operation, and an observed market or economic state.  The inventory is a project-specific synthesis than a one-to-one reproduction of the earlier cited ontologies.  Its purpose is to prevent cases such as an `entity's promise to provide funding' from being treated as though `funding had already been provided'.

\paragraph{Financial-action family.}
The 21-valued family records the broad domain function of the event.  It
contains communicative and policy-process families, public financial and
regulatory interventions, private organizational and market operations,
legal/enforcement events, and observed economic outcomes.  The family layer
is coarser than a mechanism but more finance-specific than a general event
ontology.  It is a project synthesis informed by CAMEO, policy-tool
inventories, and recurring predicates in the corpus. The distribution ratio across all policies can be found in Figure~\ref{fig:policy_family} from the appendix.

\paragraph{Financial-action subtype.}
The 99-valued subtype records the most specific supported mechanism, such as a
rate change, liquidity facility, capital injection, asset transaction,
disclosure, merger, default, investigation, or market movement.  Subtype
values are not globally interchangeable: each is licensed only under one or
more specified families.  The family--subtype constraint prevents impossible
combinations and makes the subtype suitable as an optional, higher-resolution
prediction target.
The distribution ratio across all policies can be found in Figure~\ref{fig:eventframe} from the appendix.

\paragraph{Six external crosswalk systems}

The external systems are views rather than rival ground truths.  A crosswalk
is populated only when its eligibility conditions are satisfied; otherwise
\verb|not_applicable| is a meaningful structural value rather than a low
confidence fallback.

\begin{enumerate}
  \item \textbf{CAMEO/GDELT.}  The CAMEO view records an official root,
  optional child code, and derived verbal/material cooperation/conflict
  quadrant.  It is appropriate for actor-to-actor political interactions
  \citep{schrodt2012cameo}, but not every financial state change.

  \item \textbf{Federal Reserve policy tools.}  This view identifies a
  recognized monetary-policy implementation instrument or facility using a
  24-tool inventory plus \verb|not_applicable|
  \citep{federalreservepolicytools}.  It should not be inferred merely because
  the Federal Reserve is mentioned.

  \item \textbf{ACE/ERE-style events.}  A general event view supplies an event
  type and subtype for events licensed by the ACE/ERE-style inventory
  \citep{walker2006ace}.  It preserves interoperability with general event
  extraction while leaving finance-specific distinctions to the primary
  layers.

  \item \textbf{Monetary-policy event class.}  This view distinguishes target
  actions, policy-path or forward-guidance information, implementation tools,
  and observed outcomes, following the empirical distinction between policy
  actions and information about the expected policy path
  \citep{gurkaynak2005actions}.

  \item \textbf{iMaPP macroprudential tools.}  This view maps eligible actions
  to the IMF's 17 macroprudential instrument groups plus
  \verb|not_applicable| \citep{alam2019imapp}.  It is activated by the
  instrument described, not by generic claims that a policy is
  ``macroprudential.''

  \item \textbf{IMF crisis response.}  This view captures eight systemic
  banking-crisis response forms plus \verb|not_applicable|, including
  liquidity support, guarantees, recapitalization, asset management, and
  resolution measures \citep{laeven2018crises}.
\end{enumerate}

\subsection{Validation Results}
\paragraph{Actor Validation} 
% After wikidata conflicts from extracted agents, 
% The actor classification  Out of all (Appendix~\ref{app:ergoproxy})
Table~\ref{tab:actor-resolution-coverage} in appendix summarizes actor identity resolution and its coverage across the corpus.In all organisational classifications, manual verification identified only two incorrect actor assignments among the 4530 organisation strings (including textual), corresponding to an observed accuracy of 99\%. The two errors were corrected in the final database.

\paragraph{Action Validation}
Table~\ref{tab:interaction-mode-agreement} indicates the amount of times where the \texttt{Sol} and the annotator agreed with the labels from \texttt{GPT-5.5}. The remaining conflicts were adjucated by the human again.

Table~\ref{tab:v1-database-corrections} showcases the amount of times \texttt{GPT-5.5} classifications flouted the deterministic rule and had to undergo auto-repair.

\begin{table}[t]
\centering
\caption{Independent-reviewer agreement for interaction mode.}
\label{tab:interaction-mode-agreement}
\small
\begin{tabular}{lrr}
\hline
Comparison & Agree & Rate \\
\hline
Sol--Human, initial & 2255/2522 & 89.4\% \\
Human--final resolved & 2437/2522 & 96.6\% \\
Sol--final resolved & 2334/2522 & 92.5\% \\
\hline
\multicolumn{3}{l}{Final label among 267 initial conflicts} \\
Matched Human only & 182 & 68.2\% \\
Matched Sol only & 79 & 29.6\% \\
Matched neither & 6 & 2.2\% \\
\hline
\end{tabular}
\end{table}

\begin{table}[t]
\centering
\caption{Deterministic Database Corrections.}
\label{tab:v1-database-corrections}
\small
\begin{tabular}{lrr}
\hline
Correction class & Rows & DB share \\
\hline
Resolved interaction mode & 1204 & 3.8\% \\
Hard frame constraints & 1464 & 4.7\% \\
Legacy type synchronization & 1674 & 5.3\% \\
\hline
Any corrected action & 3814 & 12.1\% \\
\hline
\end{tabular}
% \parbox{\columnwidth} this cause error
\end{table}

\section{Case Studies}
\label{sec:dataset_validation}
We conduct a focused case study validation of two policies; the September 2008 short-selling ban (Policy 5) \cite{shortsale_a} and the move to decimalization in 2001 (Policy 10) \cite{decimalization_a}. The short-selling ban involves a tightly defined regulatory intervention within a short crisis-period event window, a clearly affected financial sector, and clearly defined timelines and outcomes found in three highly cited finance papers. There were lots of news and contrasting opinions about this policy, among broader discussions of the 2008 financial crisis. Meanwhile, decimalization was implemented almost a year after it was first announced, with low news coverage and no obvious effects on the stock market. Validation on these two policies demonstrates the fidelity of the dataset across both a complex dynamic within a tight window, and a sparse information landscape across a longer time period.

\begin{figure}[h]
    \centering
    \includegraphics[width=0.75\linewidth]{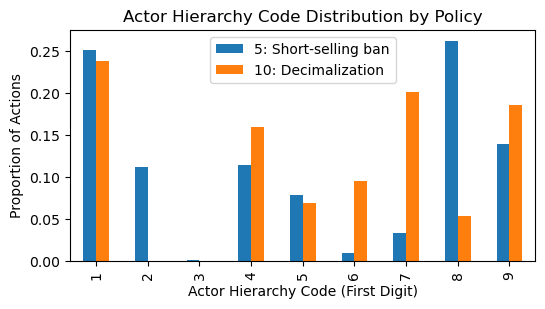}
    \caption{The proportion of actions done by different types of actors for each policy.}
    \label{fig:case_study_action_counts}
\end{figure}

Policy 5 has 62 different news articles, from which 639 actions have been extracted. Policy 10 has 21 news articles and 191 actions. Using the broadest categorisation of the actors (1st digit hierarchy code), the distribution of actors is shown in Figure \ref{fig:case_study_action_counts}. As expected, both policies have a high proportion of actors in category 1 (United States Official Bodies), which include entities such as the US Federal government and SEC, who create and announce the financial policies. Policy 5 has a large peak at category 8 (Banks, Insurance and Brokerages) since the short-selling ban targeted financial stocks. Policy 10 does not have any actors in categories 2 and 3 (Non-US Countries, and Global Bodies) since the decimalization policy only targeted the US, whereas short-selling bans were also implemented in the UK and discussed globally during the 2008 crisis.

\subsection{Policy 5: Short-selling ban}
\label{sec:Short-selling_ban}

Research published in top finance journals outlined the timeline of the 2008 short-selling ban and found that while the ban was meant to stabilize prices during the crisis, it actually severely degraded market quality by further reducing liquidity measured through widened bid-ask spreads \cite{shortsale_a, shortsale_b, shortsale_c}. Each of the events in the paper's timeline is accurately reflected in PAWS. On September 18, 2008, the United Kingdom implemented a temporary ban on short-selling of 32 financial stocks, and the US followed with a temporary ban on all short sales in 797 financial stocks. This is accurately reflected by 17 actions in the \texttt{ActionFrame} table from Sept 18-19, by actors accurately categorized as the SEC, the US federal government, and the UK. On September 22, the major exchanges announced additions to the list of banned stocks, (action 777), then on October 2, the ban was extended (action 962) and finally ended on October 8 (action 1053). A detailed list of the exact \texttt{action\_id}s is provided in Appendix \ref{app:case_study_detailed}. 

\begin{figure}[h]
    \centering
    \includegraphics[width=0.8\linewidth]{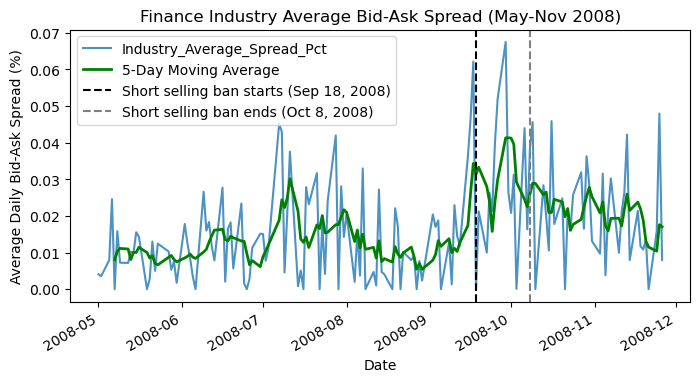}
    \caption{The bid-ask spread increased during and after the short-selling ban.}
    \label{fig:case_study_shortsale_liquidity}
\end{figure}

Since the link is not obvious, there are no actions that explicitly blame the ban as a cause of the decrease in liquidity, but there is one action which discusses the liquidity crunch after the ban. The effect is also seen in numerical data. We use the Corwin-Schultz method \cite{bid-ask_spread_calculation} to estimate bid-ask spreads using publicly available daily high and low prices from Yahoo finance across 11 representative finance stocks. Figure \ref{fig:case_study_shortsale_liquidity} shows that liquidity degraded (bid-ask spreads increased) after the short-selling ban was implemented, and persists even after the ban ended.

\begin{figure}[h]
    \centering
    \includegraphics[width=0.8\linewidth]{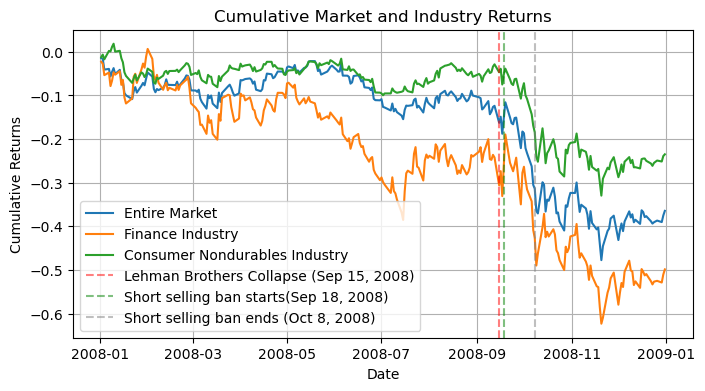}
    \caption{Average cumulative returns of the market, the finance industry, and a non-finance industry around the 2008 financial crisis.}
    \label{fig:case_study_shortsale_cum_returns}
\end{figure}

\begin{figure}[h]
    \centering
    \includegraphics[width=0.8\linewidth]{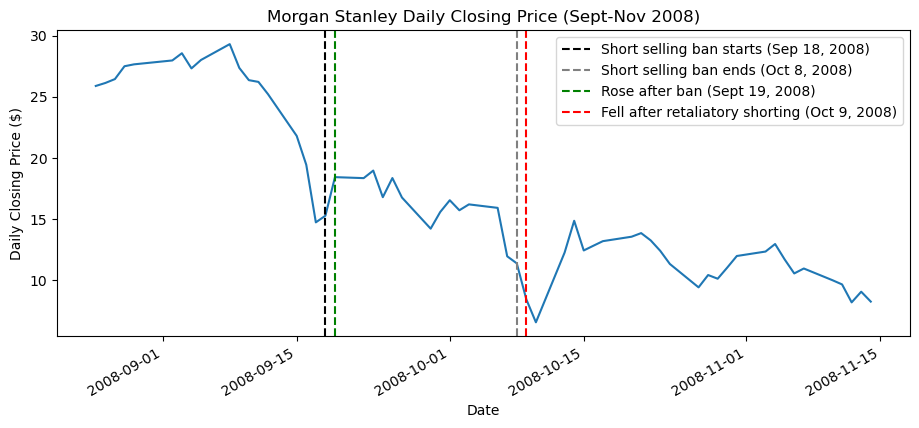}
    \caption{Morgan Stanley's stock price from September to November 2008. The dates of large stock price movements extracted from the news are highlighted and shown to correspond with the numerical data.}
    \label{fig:case_study_shortsale_ms}
\end{figure}

Beyond the ground truths provided in literature, PAWS provides information on the complex actions of other actors during the same period. By manually looking through the rest of the 639 actions for Policy 5, we find some interesting insights and dynamics. Non-finance stock seemed to experience some positive returns, as opposed to bank stocks. Figure \ref{fig:case_study_shortsale_cum_returns} shows that the market and industry returns provided in PAWS also reflect that though the entire market was plunging, the finance industry was hit harder. We also find many instances of bank representatives and other market participants blaming short-sellers for the downfall of several large banks and the overall market. However, there were also dissenters opposing the short-selling bans. A particularly interesting story was found in three actions, where a representative of Morgan Stanley lobbied for the short-sale restriction, angering many hedge funds, and led to retaliatory short-selling attacks after the ban was lifted. This is also reflected in the numerical stock price data shown in Figure \ref{fig:case_study_shortsale_ms}.

\subsection{Policy 10: Decimalization}
\label{sec:Decimalization}

\citeauthor{decimalization_a} (\citeyear{decimalization_a}) found that decimalization substantially reduced quoted bid-ask spreads, reducing trade execution costs for small trades. This effect was accurately predicted in action 4388 by the Owen Graduate School of Management, which is accurately classified as a University (hierarchical code 7210). The SEC order to start phasing in decimalization, the time when the NYSE completed the implementation, the news of delays, and when the Nasdaq exchange also completed implementation, are all accurately reflected on the correct dates with the right hierarchical classifications.

Other interesting actions include suspicions and denial of insiders profiting from the previous fractional pricing (1/16th of a dollar), and explanations of the benefits decimal pricing will bring to investors from both academic institutions and government bodies pushing for decimalization.

These two case studies show that PAWS is able to accurately extract dated actions from sparse news for policies more than 20 years ago, as well as accurately reflect the complex and dynamic actions taken by a variety of actors during a dense policy window.

\section{Simulation Study}
\label{sec:simulation-results}

We show an example of how the PAWS dataset can be used to study multi-agent behaviour and interaction by showcasing how stakeholder actions can be predicted from the policy-agent-day panel. 

\paragraph{Policy 5 full-process replay} 

\begin{table}[t]
\centering
\small
\setlength{\tabcolsep}{3pt}
\renewcommand{\arraystretch}{1.08}
\begin{tabular}{l r r r r r}
\toprule
% \textbf{Model} & \textbf{N} & \textbf{Acc.} & \textbf{Macro-F1} & \textbf{Active-F1} & \textbf{Active Recall} \\
\textbf{Model} & \textbf{N} & \textbf{Acc.} & \textbf{Macro-} & \textbf{Active-} & \textbf{Active} \\
 & & & \textbf{F1} & \textbf{F1} & \textbf{Recall} \\
\midrule
Majority & 8341 & .980 & .330 & .000 & .000 \\
Transition & 8341 & .979 & .330 & .012 & .006 \\
Threshold ABM & 8341 & .969 & .349 & .080 & .067 \\
Hybrid ABM & 8341 & .947 & .375 & .095 & .139 \\
Structured classifier & 8341 & .877 & .356 & .112 & .382 \\
\texttt{GPT-4o} replay & 50 & .240 & .205 & .718 & .560 \\
\bottomrule
\end{tabular}
\caption{Minimal multi-policy replay results from \texttt{minimal\_first\_study\_outputs\_llm}. The \texttt{GPT-4o} replay is bounded to a 50-row sample and should not be compared to full-panel models by raw accuracy.}
\label{tab:minimal-replay-results}
\end{table}

\begin{figure}[t]
    \centering
    \includegraphics[width=0.8\linewidth]{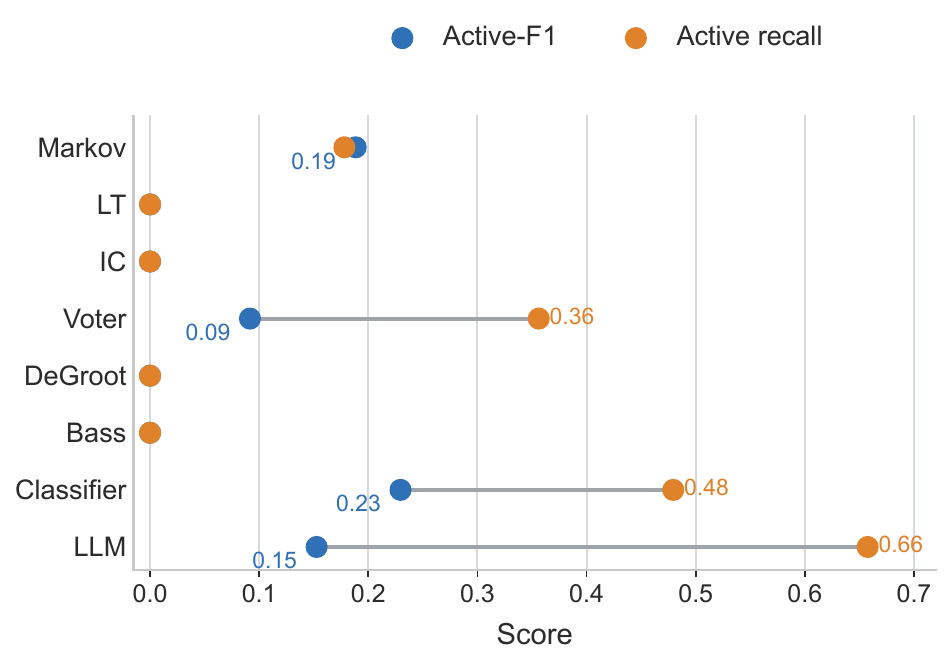}
    \caption{Policy 5 full-process replay action recovery. The structured classifier offers the best active-F1, while the \texttt{GPT-4o} LLM agent obtains the highest active recall but over-triggers relative to the observed action rate.}
    \label{fig:policy5_action_recovery}
\end{figure}

No-action diffusion baselines reach high accuracy (.939) but zero active-F1. The \texttt{GPT-4o} LLM-agent variant reaches .824 active recall but only .122 active-F1, while the structured classifier is more conservative, with .069 active-F1 and .176 active recall. Figure~\ref{fig:policy5_action_recovery} reports the full-process replay for the September 2008 short-selling ban. The no-action diffusion baselines obtain high accuracy because the true active rate is only 8.7\%, but fail to recover active stakeholder responses. 
More information of the replay can be found in Appendix~\ref{app:simulation_diagnostics}.

\section{Conclusions and Future Work}

We introduced PAWS, a policy-centered dataset that connects verified financial-policy episodes with dated news evidence, source-grounded stakeholder actions, normalized organizations, structured event frames, and daily market context.

The validation and case-study results support the usefulness of this design while also revealing the difficulty of the task. Independent AI and human reviewers achieved 89.4\% initial agreement on interaction-mode labels, after which disagreements were adjudicated. The case studies recover documented intervention timelines and associated market patterns under substantially different levels of news coverage. The replay study additionally demonstrates that aggregate accuracy is misleading in sparse policy-agent-day panels: models can achieve high accuracy by predicting \emph{no action} while failing to recover the stakeholder responses that matter. Meaningful evaluation must therefore emphasize active-action recall, timing, and the fidelity of resulting response cascade.

PAWS should be interpreted as a curated observational and replay resource than a causal estimate of policy effects. Future work should expand the range of policies and jurisdictions, strengthen human validation, improve agent calibration, and incorporate higher-fidelity market mechanisms and counterfactual evaluation. These extensions would support more rigorous studies of selective agent influence, policy-response dynamics, and the expected and unexpected consequences of financial interventions.

\section*{Acknowledgements}
This research is supported by Tsinghua University Initiative Scientific Research Program.

\bibliography{custom}

\newpage
\onecolumn 
\appendix
\providecommand{\code}[1]{\path{#1}}
\providecommand{\NA}{\textsc{n/a}}

\section{Prompt used for Policy Identification}~\label{appn:prompts}
\begin{tcolorbox}[title=Prompt for Policy Identification, colback=gray!10, colframe=black!50, boxrule=0.5mm]
\begin{verbatim}
You are a renown tenured finance professor at a prestigious university. You
have been researching various financial topics for decades, and are an expert
at identifying changes in financial policies that had large impacts on the
financial market. You often leverage these as "exogenous shocks" to
investigate causality in your research. 

Please provide a list of policy changes from that have made large impacts on
the US financial markets, no matter if they were simply short-term effects
or long-term effects. For each policy, provide the reason the policy was
implemented and the intended outcomes of the policy. Also provide any
published/working papers from reputable sources (i.e. published in top
finance journals or written by authors from prestigious universities) that
studied the effects of the policy and summarize their established impact(s)
on the financial markets. For policies that were not used in any research
papers, provide your own summary on their impacts on the financial markets,
and relate to the intended outcomes of the policy. Try to focus on impacts
that can be measured using stock market returns.

Be as comprehensive as possible, covering as many policies as possible, but
keep the summaries of the impacts concise and succinct.

Output your response in a single text block in the following JSON format:
{
  "policy_name": "the name of the policy",
  "announcement_date": "YYYY-MM-DD",
  "implementation_date": "YYYY-MM-DD",
  "description": "a brief description of what the policy does and why it was
  implemented",
  "intended_outcomes": "a summary of the intended outcomes of the policy",
  "research_papers": [
    {
      "title": "title of the research paper",
      "authors": ["name1", "name2"],
      "link": "link to the paper",
      "findings": "a concise summary of what the authors found to be the
      impact of the policy"
    }
  ]
  "llm_analysis": "your own analysis of the impact of the policy"
}
\end{verbatim}
\end{tcolorbox}

\clearpage
\section{Policy Details}~\label{appn:PolicyInfo}

In order of announcement/implementation date, numbered by \texttt{policy\_id}.

\subsection{10: Decimalization}
On 9 April 2001, US markets switched from fractional pricing (1/16th of a dollar) to decimal pricing (pennies), reducing the minimum tick size. This was done to improve market efficiency by allowing prices to more accurately reflect stock value (1/100th instead of 1/16th precision). Research documented that quoted bid-ask spreads fell dramatically, reducing transaction costs and increasing liquidity for small trades \citeapp{decimalization_a}.

\subsection{12: JGTRRA}
Congress approved the Jobs and Growth Tax Relief Reconciliation Act of 2003 on 23 May 2003, and it was signed into law on 28 May 2003. The act reduced the maximum tax rate on capital gains from 20\% to 15\% and the maximum tax rate on dividends from 38.1\% to 15\%. Literature found a 20\% increase in dividend payments by nonfinancial, nonutility publicly traded corporations following the tax cut \citeapp{JGTRRA}.

\subsection{8: Term Auction Facility (TAF)}
Announced on 12 December 2007, the Term Auction Facility's (TAF) primary objective was to provide emergency liquidity, reduce the liquidity risk premium in the interbank market, and consequently lower the significantly elevated London interbank offered rate (Libor).
Literature documented mixed results, but seems to imply that the policy effectively relieved the liquidity strains and brought down the Libor \citeapp{mcandrews_sarkar_wang_2015}.

\subsection{5: Short-Selling Bans}
On 18 September 2008, the UK and US banned short selling on financial stocks to stabilize prices, and this ban was effective until 8 October 2008 for the US market. However, this policy was largely counterproductive, severly degrading market quality (spreads widened, price impact increased, and intraday volatility rose)\citeapp{shortsale_a}, slowed price discovery\citeapp{shortsale_b}, and dramatically increased bid-ask spreads\citeapp{shortsale_c}.

\subsection{7: Commercial Paper Funding Facility (CPFF)}
On 7 October 2008, the Federal Reserve announced the Commercial Paper Funding Facility (CPFF) to provide a liquidity backstop; to ensure that firms have enough liquidity to meet their operational needs and not go bankrupt.
The literature found that firms eligible for the CPFF were able to mitigate financing disruptions and led to an increase in profitability and short-term earnings forecasts, with spillover effects also leading to an increase in general public liquidity \citeapp{gao_yun_2012}.

\subsection{6: Term Asset-Backed Securities Loan Facility (TALF)}
On 25 November 2008, the Term Asset-Backed Securities Loan Facility (TALF) was announced. The TALF was intended to increase demand for asset-backed securities (ABS) and encourage more lending to households and small businesses.
The literature document that TALF played a critical role in preventing the shutdown of lending to consumers and small businesses \citeapp{CAMPBELL2011518_TALF}.

\subsection{18: The Durbin Amendment}
Announced on 29 June 2001 and implemented on 1 October 2011, the Durbin Amendment capped interchange fees charged to merchants for processing debit card transactions to reduce retailer costs with the hope that those savings would be passed to consumers as lower prices. Research found that banks offset \$6.5 billion in losses by increasing maintenance fees and reducing free checking availability \citeapp{MUKHARLYAMOV2025104094}, hence consumers ultimately paid more through increased banking fees than they saved at the register.

\subsection{23: Say-on-Pay Mandates}
Announced on 21 July 2010 and implemented on 1 April 2011, the Say-on-Pay Mandates required public companies to hold advisory shareholder votes on executive compensation in an effort to curb excessive compensation and align executive incentives with shareholder interests. Literature found that implementation in April 2011 led to a 4.6\% increase in firm market value and improved long-term operational productivity \citeapp{10.1093/rof/rfv056}.

\subsection{21: Budget Control Act / S\&P Sovereign Downgrade}
Announced on 2 August 2011 and implemented on 5 August 2011, the Budget Control Act (BCA) was enacted to resolve debt ceiling crisis while the S\&P subsequently downgraded U.S. credit to AA+. The BCA aimed for deficit reduction and the downgrade reflected political and fiscal governance concerns. Research found that when the US Federal government repeatedly reached the legally binding debt limits, the financial market only initially charged a small default risk premium to the Treasury securities but no significant overall price effect, suggesting that the financial market gradually perceived the budget standoffs as "the boy who cried wolf" \citeapp{LIU20091464}. 

\subsection{13: Operation Twist (Maturity Extension Program)}
Announced on 21 September 2011 and implemented on 1 October 2011, Operation Twist was a sterilized monetary initiative where the Fed sold \$400 billion in short-term Treasuries to purchase long-term Treasuries, with the aim to flatten the yield curve, reduce long-term interest rates, and encourage portfolio rebalancing into riskier assets without expanding the balance sheet. Literature found a significant 15 basis point decline in long-term Treasury yields and a 2-4 basis point decline in corporate bond yields \citeapp{operation_twist}, indicating that the program was successful.

\subsection{20: G-SIFI Designations}
On 1 November 2011, specific institutions were designated as 'too big to fail,' subjecting them to capital surcharges. This was done in an effort to internalize negative externalities and reduce the probability of systemic failures. Research found that designated entities faced higher capital costs, leading to lower stock valuations compared to non-designated peers \citeapp{scott2012interconnectedness}, creating a 'regulatory discount' on major bank stocks and forcing a retrenchment from capital-intensive activities.

\subsection{16: Dollar Liquidity Swap Line Rate Reduction}
Announced on 30 November 2011 and implemented on 5 December 2011, the Fed and other central banks lowered the pricing on dollar liquidity swap arrangements by 50 basis points, in hopes of relieving offshore dollar funding stress and preventing fire sales of dollar assets by European banks. Research found that it reduced Covered Interest Parity (CIP) deviations and stabilized cross-border investment flows \citeapp{10.1093/restud/rdab074}, acting as a crucial global lender-of-last-resort mechanism, and removing systemic tail risk during the European debt crisis.

\subsection{22: Jumpstart Our Business Startups (JOBS) Act}
On 5 April 2012, the JOBS Act eased securities regulations and auditing requirements for 'Emerging Growth Companies' (EGCs) to encourage IPOs and lower capital formation barriers for high-growth startups. Literature found that it boosted IPO volume by 25\%, particularly in biotech, but led to significantly higher IPO underpricing \citeapp{DAMBRA2015121}.

\subsection{19: U.S. Implementation of Basel III}
Announced on 7 June 2012 but implemented a year later on 1 July 2013, the US increased the quantity and quality of required Common Equity Tier 1 (CET1) capital for banks to ensure banks maintain high-quality capital buffers to absorb losses during economic stress. Research found that after the announcement, banks had lowered return on equity and accelerated migration of credit provision to the shadow banking sector, moving outside the regulatory perimeter \citeapp{fritsch_siedlarek_2022}.

\subsection{15: Quantitative Easing 3 (Open-Ended LSAP)}
On 13 September 2012, the US made open-ended monthly purchases of \$40 billion in agency mortgage-backed securities (MBS) in efforts to compress mortgage rates, stimulate housing, and create a wealth effect through higher equity prices. Research found that it increased global equity prices, lowered the VIX, and induced massive capital inflows into emerging markets \citeapp{EICHENGREEN20151}.

\subsection{14: State-Contingent Forward Guidance (The Evans Rule)}
Announced and implemented on 12 December 2012, the Evans Rule replaced calendar-based guidance with explicit economic thresholds: 6.5\% unemployment and 2.5\% inflation, to eliminate interest rate uncertainty and provide an 'automatic stabilizer' for market psychology. Research found that it functioned as an Odyssean commitment mechanism, recalibrating futures contracts and private forecasts \citeapp{evans_rule}, ensuring monetary accommodation would persist through economic weakness.

\subsection{24: Unconventional Monetary Policy Tapering Announcement (Taper Tantrum)}
On 22 May 2013, Federal Reserve Chairman Ben Bernanke announced the Fed's intention to begin scaling back its \$85 billion-a-month bond-buying program (Quantitative Easing 3). The actual tapering of asset purchases began in January 2014. This was to normalize U.S. monetary policy, gracefully exit the unprecedented quantitative easing measures implemented post-2008, and prevent the U.S. economy from overheating as macroeconomic indicators showed sustained recovery. Literature found that the tapering announcement functioned as a major monetary shock that dramatically increased required risk compensation, triggering massive capital outflows from emerging markets, disproportionately depressing emerging market equities and currencies compared to physical capital flows \citeapp{10.1093/rfs/hhaa044}.

\subsection{25: Title II of the JOBS Act (General Solicitation)}
Announced on 10 July 2013 and implemented 23 September 2013, this was a structural deregulatory action by the SEC that eliminated the long-standing ban on 'general solicitation and general advertising' for private placements conducted under Rule 506(c), provided all purchasers are verified accredited investors. The intention was to democratize private capital formation, allowing startups and small businesses to utilize the internet and public media to reach a broader base of accredited investors, thereby expanding capital access and lowering early-stage costs. Literature found that after accounting for selection bias, issuers using general solicitation are less likely to succeed in raising capital, obtaining venture capital funding, or exiting via IPO \citeapp{Agrawal_Lim_2026}. For the public stock market, this dynamic reduced the average quality of incoming small-cap companies, characterized by higher return volatility and an increased rate of exchange delistings.

\subsection{26: The Volcker Rule (Dodd-Frank Act Section 619)}
Announced on 10 December 2013 and implemented on 1 April 2014, this rule restricts U.S. depository institutions from making speculative investments that do not benefit their customers. It explicitly prohibits commercial banks from engaging in proprietary trading and severely limits their investments in hedge funds and private equity funds. Research found that the Volcker Rule decreased corporate bond liquidity during times of stress, with dealers affected by the rule curtailing their market-making activities, and non-affected dealers not fully offseting this loss, leading to significantly increased illiquidity when bonds were downgraded \citeapp{BAO201895}.

\subsection{27: 2014 SEC Money Market Fund Reform}
Announced 23 July 2014 and implemented 14 October 2016, the SEC adopted structural reforms requiring institutional prime money market funds (MMFs) to float their net asset value (NAV) rather than maintaining a stable \$1.00 NAV. It also provided fund boards the tools to impose liquidity fees and redemption gates if a fund's weekly liquid assets fell below regulatory thresholds. Research demonstrated that the regulation reduced the 'money-likeness' of prime MMFs by increasing their information sensitivity, with investors shifting over \$1 trillion from prime to government MMFs, revealing a significant 20-30 basis point premium that investors are willing to pay for money-like stability \citeapp{CIPRIANI2021250}.

\subsection{28: Department of Labor (DOL) Fiduciary Rule}
Announced on 14 April 2015, but never implemented, the rule expanded the definition of 'investment advice fiduciary' under ERISA, legally requiring advisors and broker-dealers to act in the best interests of their clients on retirement accounts. Research revealed that variable annuity sales were highly sensitive to broker commissions, with sales of high-expense annuities plummeting by 52\% post-rule announcement, shifting capital toward lower-cost products and improving investor welfare overall \citeapp{10.1093/rfs/hhac047}. Literature also demonstrated that ETF prices and the equity valuations of ETF sponsors reacted with significant positive abnormal returns following the DOL rule announcements \citeapp{DOL_b}.

\subsection{29: Repeal of the U.S. Crude Oil Export Ban}
Announced on 15 December 2015 and implemented 2 days later, the US legislative repealed the 1970s-era statutory prohibition on exporting domestically produced crude oil to global markets, in an effort to relieve transportation and refining bottlenecks caused by the shale boom, close the pricing spread between WTI (US oil price) and Brent crude (global oil price), and stimulate domestic energy exploration and production. Literature found that repealing the export ban led to increased global trade and long-term efficiency in the global oil market \citeapp{repec:fip:fedker:00053}, stabilized oil prices \citeapp{10.3389/fphy.2020.551501}, and relieved artificially low light oil prices in the U.S., allowing exports to expand and correcting inefficient production mandates on U.S. refineries \citeapp{LANGER2016258}.

\subsection{30: Federal Reserve Zero Lower Bound (ZLB) Liftoff}
On 16 December 2015, the FOMC raised the target range for the federal funds rate from 0.00-0.25\% to 0.25-0.50\%, ending seven consecutive years of Zero Interest Rate Policy. Literature documented the abrupt disappearance of the 'pre-FOMC announcement drift'(where the stock market saw significant abnormal positive returns within 24 hours before Federal Open Market Committee (FOMC) meetings) in equity markets, a phenomenon that vanished precisely coinciding with the ZLB liftoff \citeapp{KUROV2024100446}.

\subsection{31: Treasury Department Section 385 Regulations (Anti-Inversion and Earnings Stripping Rules)}
Announced on 4 April 2016 and implemented on 21 October 2016, the IRS broad administrative was granted the authority to recharacterize certain intercompany debt instruments as equity to stop 'earnings stripping' practices during corporate inversions, in an effort to halt corporate tax inversions (specifically targeting the pending Pfizer-Allergan mega-merger) and to protect the U.S. corporate tax base from erosion via cross-border related-party debt structuring. Literature demonstrates that U.S. firms engaged in offshore re-domiciling experience significant negative cumulative abnormal returns when the U.S. government unexpectedly tightens the tax code \citeapp{reyes_upadhyay}.

\subsection{32: The Tax Cuts and Jobs Act (TCJA)}
Announced on 20 December 2017 and implemented on 1 January 2018, the TCJA was a comprehensive overhaul of the US tax code that reduced the statutory corporate income tax rate from 35\% to 21\%, shifted the US to a territorial tax system, mandated a transition tax on accumulated foreign earnings, and limited interest expense deductibility. Literature found that domestic firms historically paying high effective tax rates experienced significant positive abnormal returns, while internationally oriented firms suffered due to the unexpected magnitude of the repatriation transition tax, and median effective tax rates fell from 31.7\% to 20.8\% \citeapp{wagner_zeckhauser_ziegler_2020}.

\subsection{33: U.S.-China Trade War Tariff (1)}
On 22 March 2018, this was the first of a series of unilateral and escalating retaliatory import tariffs levied by the US and China against each other's goods, spanning multiple tranches throughout 2018 and 2019. This was when the Trump administration issued the presidential memorandum proposing tariffs on up to \$50 billion of Chinese imports. US firms exposed to China experienced immediate and persistent stock return declines; within 2 days of the announcement date, a one standard deviation increase in a firm's share of sales to China led to a 0.48\% lower cumulative return, while a one standard deviation increase in direct inputs from China led to a 0.29\% lower cumulative return \citeapp{HUANG2023103811}. Within a 7-day window of the announcement date, the market experienced a cumulative return of -3.39\%, of which -2.62\% was directly cause by the tariff announcement, and firms exposed to the Chinese market saw an additional -0.04\% impact \citeapp{amiti2020effect}. At 40 days after the announcement, a 10 percentage-point increase in a firm's share of revenue from China is associated with a 2.3\% lower cumulative stock return, and remains significant up till 80 days post announcement \citeapp{HUANG2023103811}.

\subsection{36: Economic Growth, Regulatory Relief, and Consumer Protection Act (EGRRCPA)}
Announced on 22 May 2018 and implemented 2 days later, this legislation rolled back specific Dodd-Frank Act provisions, most notably raising the asset threshold for strict regulatory stress tests from \$50 billion to \$250 billion and exempting community banks from the Volcker Rule. This was to reduce compliance costs for small and mid-sized banks, free up capital for economic growth and consumer lending, and slow down community bank consolidations. After the EGRRCPA on 22 March 2018, small banks (under \$10 billion in assets) reacted most favorably, showing tangible subsequent growth in lending and asset growth, but no observed effects on large banks \citeapp{coffee_2022}.

\subsection{34: U.S.-China Trade War Tariff (2)}
On 17 September 2018, the second of a series of unilateral and escalating retaliatory import tariffs levied by the US and China against each other's goods, spanning multiple tranches throughout 2018 and 2019. This was the announcement of tariffs on \$200 billion of Chinese imports. This announcement had very little net impact on the market. It generated a mild negative effect on the entire market (-0.60\%) alongside a small positive differential effect (+0.26\%) for exposed firms \citeapp{amiti2020effect}, suggesting it may have been less protectionist than market participants initially anticipated.

\subsection{35: U.S.-China Trade War Tariff (3)}
On 10 May 2019, the third of a series of unilateral and escalating retaliatory import tariffs levied by the US and China against each other's goods, spanning multiple tranches throughout 2018 and 2019. The U.S. escalated the September 2018 tariffs on the \$200 billion list from 10\% to 25\%. This announcement heavily penalized China-exposed firms via a -1.36\% drop in returns (especially hurting firms with high revenue exposure to China or reliant on Chinese imports)., and simultaneously dragged down the broader market by -1.62\% \citeapp{amiti2020effect}.

\subsection{37: The September 2019 Repo Market Intervention and FOMC Policy Pivot}
On 17 September 2019, the Federal Reserve implemented emergency interventions in the repurchase agreement (repo) market to inject liquidity after overnight borrowing rates spiked above 5\%. This catalysed a pivot away from quantitative tightening and toward interest rate cuts. This was done to stabilize the Secured Overnight Financing Rate (SOFR), restore liquidity to short-term funding markets, and prevent downward pressure on the broader economy. Textual analysis reveals a causal effect of the stock market on Fed policy, where negative stock returns are strongly correlated with Fed growth expectation downgrades, which dictate accommodating monetary policy (the 'Fed Put', the market's belief that the central bank will step in to stop severe stock market drops) \citeapp{cieslak2020economics}.

\subsection{38: Final Rule Implementing Amendments to the Volcker Rule (Volcker 2.0 / 2.1)}
Announced on 14 November 2019 and effective 1 January 2020, these regulatory revisions simplified Volcker Rule compliance by creating an objective safe harbor (instruments held over 60 days are presumed not to be proprietary trading) and removing highly subjective intent tests. Research found that the original Volcker Rule structurally impaired liquidity and the 2019/2020 revisions were enacted to successfully reverse this illiquidity \citeapp{BAO201895}. 

\subsection{39: Federal Reserve Emergency Rate Cuts and Corporate Credit Facilities (PMCCF/SMCCF)}
Announced on 23 March 2020 and implemented on 16 June 2020, these were unprecedented emergency monetary policy in response to COVID-19, cutting rates to zero and establishing facilities that authorized the central bank to directly purchase corporate debt and corporate bond ETFs. Research found that the facility announcements reduced credit spreads on eligible bonds by 70 basis points within 10 days and restored the normal yield curve, while actual physical purchases after the implementation date lowered eligible bond spreads by an additional 11 basis points immediately, with persistent effects for low-quality bonds \citeapp{GILCHRIST2024103573}. The announcement also triggered an immediate, dramatic stabilization of the corporate bond market by substantially lowering the default risk premium, with average secondary market credit spreads droppinf by over 90 basis points within three days, and strictly eligible investment-grade bonds experiencing an additional differential decline of 52 basis points \citeapp{BOYARCHENKO2022695}. The announcement effectively functioned as a "buyer of last resort," significantly reducing fire-sale incentives by narrowing bid-ask spreads for highly-rated and BBB eligible bonds by more than 45 cents and 30 cents, respectively \citeapp{BOYARCHENKO2022695}.

\subsection{40: Coronavirus Aid, Relief, and Economic Security (CARES) Act}
Announced on 25 March 2020 and implemented on 13 April 2020, the act enacted a \$2.2 trillion emergency fiscal stimulus package addressing the COVID-19 lockdowns, which included direct Economic Impact Payments (stimulus checks) sent to US households. The arrival of stimulus checks on April 13, 2020, triggered a massive spike in retail trading targeting low-priced, high-volatility stocks, with retail-concentrated portfolios seeing 5.62\% abnormal returns immediately on the first day, growing to 10.73\% cumulative abnormal returns over 3 days and 14.74\% over three weeks \citeapp{greenwood2022stock}.

\subsection{41: American Rescue Plan Act of 2021 (Third Round Stimulus Checks / EIP3)}
Announced on 11 March 2021 and implemented the next day, the American Rescue Plan authorized a massive fiscal package that included a third round of direct Economic Impact Payments (EIP3) of up to \$1,400 per eligible individual. This constituted a sudden wealth shock distributing over \$391 billion to U.S. households. Research found that while the first two rounds of stimulus checks generated massive abnormal returns in retail-dominated stock portfolios, the third round of stimulus produced no significant positive portfolio return effects, and point estimates were insignificantly negative \citeapp{greenwood2022stock}.

\subsection{42: CHIPS and Science Act}
Announced on 18 May 2021 and implemented on 9 August 2022, this was a legislative package committing \$52.7 billion in direct federal funding, subsidies, and investment tax credits to revitalize domestic semiconductor research, design, and fabrication, in an effort to reverse the decline in U.S. semiconductor manufacturing capacity, reduce reliance on foreign supply chains, stimulate private capital expenditure, and create high-paying domestic manufacturing jobs. Research revealed significant abnormal stock returns for semiconductor firms on May 19, 2021, precisely when the precursor bill was introduced, but zero abnormal returns upon the actual passage and signing of the Act in 2022\citeapp{NBERw34625}. The act generated approximately 15,000 direct jobs \citeapp{NBERw34625}.

\subsection{43: Infrastructure Investment and Jobs Act (IIJA)}
Announced on 10 August 2021 and implemented on 15 November 2021, this was a generational infrastructure package authorizing approximately \$1.2 trillion in total spending, including \$550 billion in newly allocated federal investments across transportation, broadband, water systems, and power grids. Analyses on previous similar 'Buy American' provisions show that these legislations significantly inflated the cost of federal procurement, and macroeconomic models show that the aggregate impact on the Fed funds rate and 10-year Treasury yields is quantitatively small due to the spending being spread over a decade \citeapp{NBERw32953}.

\section{Alternative Policy Identification LLM}~\label{appn:traderLLM}
PAWS uses an LLM acting as a tenured finance professor to identify impactful policies by leveraging the prevalent use of policies as causal identification strategies in finance research. To examine if a finance research background is necessary, we test an alternative LLM for identifying policies; one that acts as a renowned manager at a top U.S. hedge fund with decades of experience.

As expected, this alternative policy identification LLM frequently hallucinates research papers and their documented policy effects. Interestingly, this alternative policy identification LLM also misidentified the announcement and implementation dates of approximately 30\% of the identified policies, and even hallucinated a policy that does not exist.

\section{Event-Frame}

\subsection{Event-Frame Forms}\label{app:framedes}
Prior to multi-layer event-frame classification, each action is normalized into three classes: \emph{Communication}, \emph{Direct Intervention}, and \emph{Observed Effect}. The normalized classification can be viewed as a legacy label, purely for reference.

The interaction-mode layer records how the event is realized, for example
\texttt{communication}, \texttt{intent\_or\_plan},
\texttt{institutional\_action}, \texttt{request\_or\_pressure},
\texttt{material\_action}, \texttt{legal\_or\_enforcement\_action},
\texttt{market\_or\_economic\_state}, or \texttt{other}. The family and subtype
layers record the domain-level financial function and more specific mechanism.
This separation is important for policy analysis: a plan to provide liquidity,
a legal authorization to provide liquidity, and an actual liquidity transfer
may share a policy function while differing in interaction mode and evidential
status.

The external crosswalks expose compatible views of the same action when
applicable: CAMEO/GDELT-style actor interactions \citeapp{schrodt2012cameo},
Federal Reserve policy tools \citeapp{federalreservepolicytools}, ACE/ERE-style
event types \citeapp{walker2006ace}, monetary-policy event classes
\citeapp{gurkaynak2005actions}, IMF iMaPP macroprudential tools
\citeapp{alam2019imapp}, and IMF crisis-response categories
\citeapp{laeven2018crises}. A crosswalk value of \texttt{not\_applicable} means
that the external ontology is not structurally licensed for the action; it does
not mean that the action is missing or invalid. 

The full categorical inventory can be found in supplementary appendix (Categorical Inventory) attached in the supplementary material.
%(Appendix~\ref{app:frameclass}).
% \input{frame_significance_table}

\subsection{Motivation} \label{app:motivation}
The first source is frame-semantic event representation.  Frame semantics
describes lexical meaning relative to a structured situation
\citeapp{fillmore1982framesemantics}; FrameNet groups lexical units by the frame
they evoke and identifies their participants \citeapp{baker1998framenet}.
PropBank and Abstract Meaning Representation similarly motivate predicate--role
representations that abstract away from some changes in syntax and surface
wording \citeapp{palmer2005propbank,banarescu2013amr}.  These traditions support
representing an actor, target, transferred resource, instrument, and amount
separately from the event's label.  Thus, \emph{paid}, \emph{provided
funding}, and \emph{injected capital} may instantiate closely related
support/transfer semantics even though their lexical realizations differ.

The second source is the separation of event content from communicative force
and factual status.  Speech-act theory distinguishes assertions, directives,
commitments, and declarations by their illocutionary point
\citeapp{searle1976illocutionary}.  ACE, FactBank, and TimeML separately motivate
event typing, modality/factuality, and temporal or lifecycle information
\citeapp{walker2006ace,sauri2009factbank,pustejovsky2003timeml}.  This distinction
underlies \verb|interaction\_mode|: announcing an intended transfer,
authorizing it, and executing it can concern the same substantive policy
function while differing in realization and commitment.

The third source is policy-instrument theory.  The tools-of-government
tradition distinguishes governing resources such as information, legal
authority, finance, and organizational capacity
\citeapp{hoodmargetts2007tools}.  Work on policy design further separates the
substantive instrument from the stage or procedure through which it is adopted
\citeapp{howlett2009governance}.  Financial-policy practice supplies more
domain-specific tool inventories: Federal Reserve policy tools, target and
path information in monetary-policy news, the IMF's iMaPP macroprudential
categories, and systemic-crisis response categories
\citeapp{federalreservepolicytools,gurkaynak2005actions,alam2019imapp,laeven2018crises}.

Finally, the intended downstream use is agent-based simulation.  BDI models
and automated planning represent actions as choices or operators available to
an actor, with interpretable conditions and effects
\citeapp{rao1995bdi,mcdermott2000planning}.  This motivates a proposed
\verb|event\_route| that distinguishes intentional \verb|agent\_action| events
from \verb|environment_state_or_outcome| events.  A fall in prices is normally
an environment transition; a central bank's sale of assets is an action that
may contribute to that transition.

\subsection{Dataset Statistics}
\paragraph{Entity}
 Starting from 13,720 unique nonblank entity strings, 7,810 (56.9\%) were assigned to an organization. Alias consolidation produced 4,317 canonical organization labels, of which 2,578 (59.7\%) were linked to distinct Wikidata QIDs.

\begin{figure}[h]
    \centering
    \includegraphics[width=0.99\linewidth]{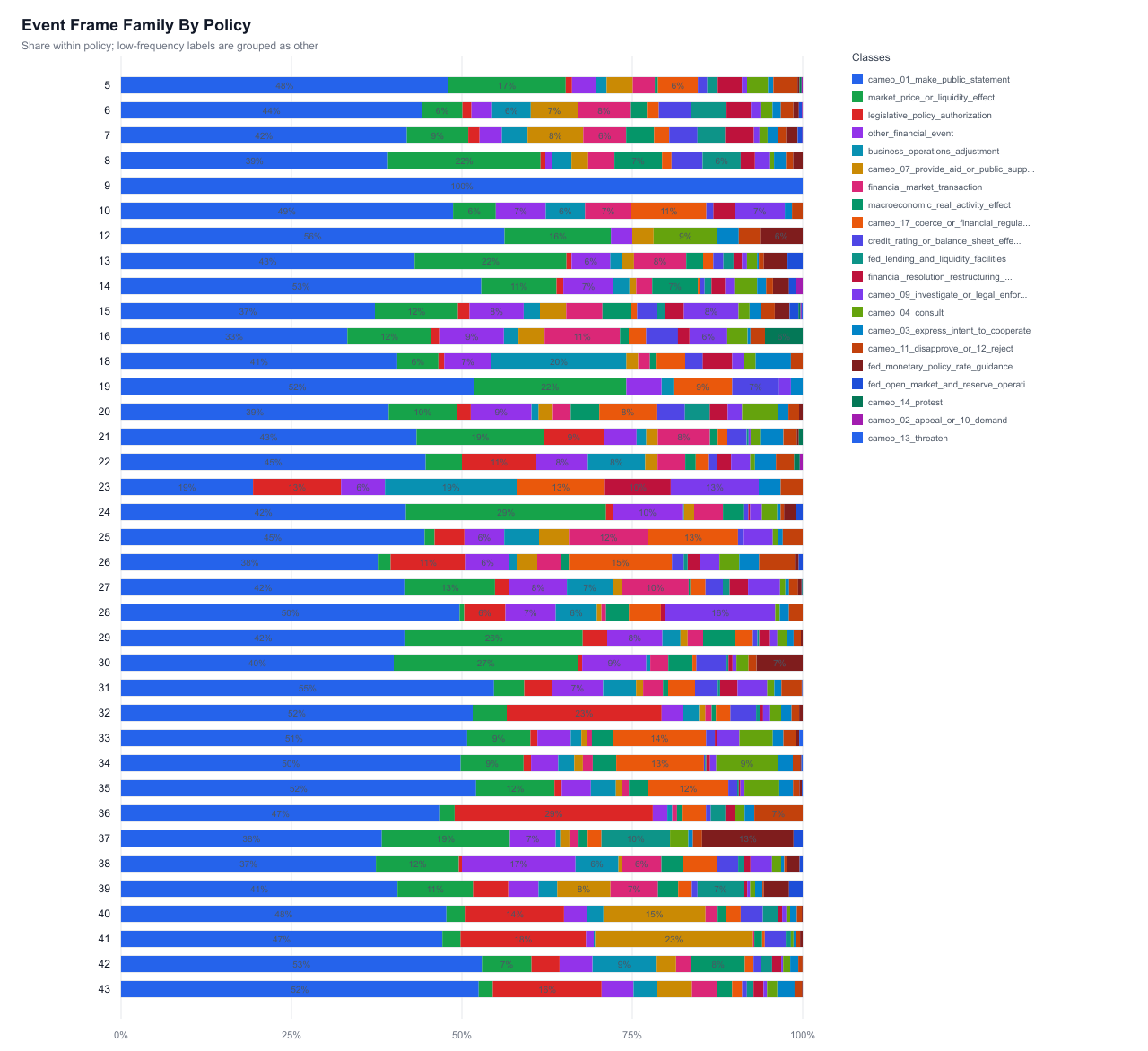}
\caption{Financial-action family composition by policy. Each bar shows the shares of the seven most frequent families, with remaining families pooled. Numeric codes correspond to the accompanying category key.}
    \label{fig:policy_family}
\end{figure}

\begin{figure}[h]
    \centering
    \includegraphics[width=0.99\linewidth]{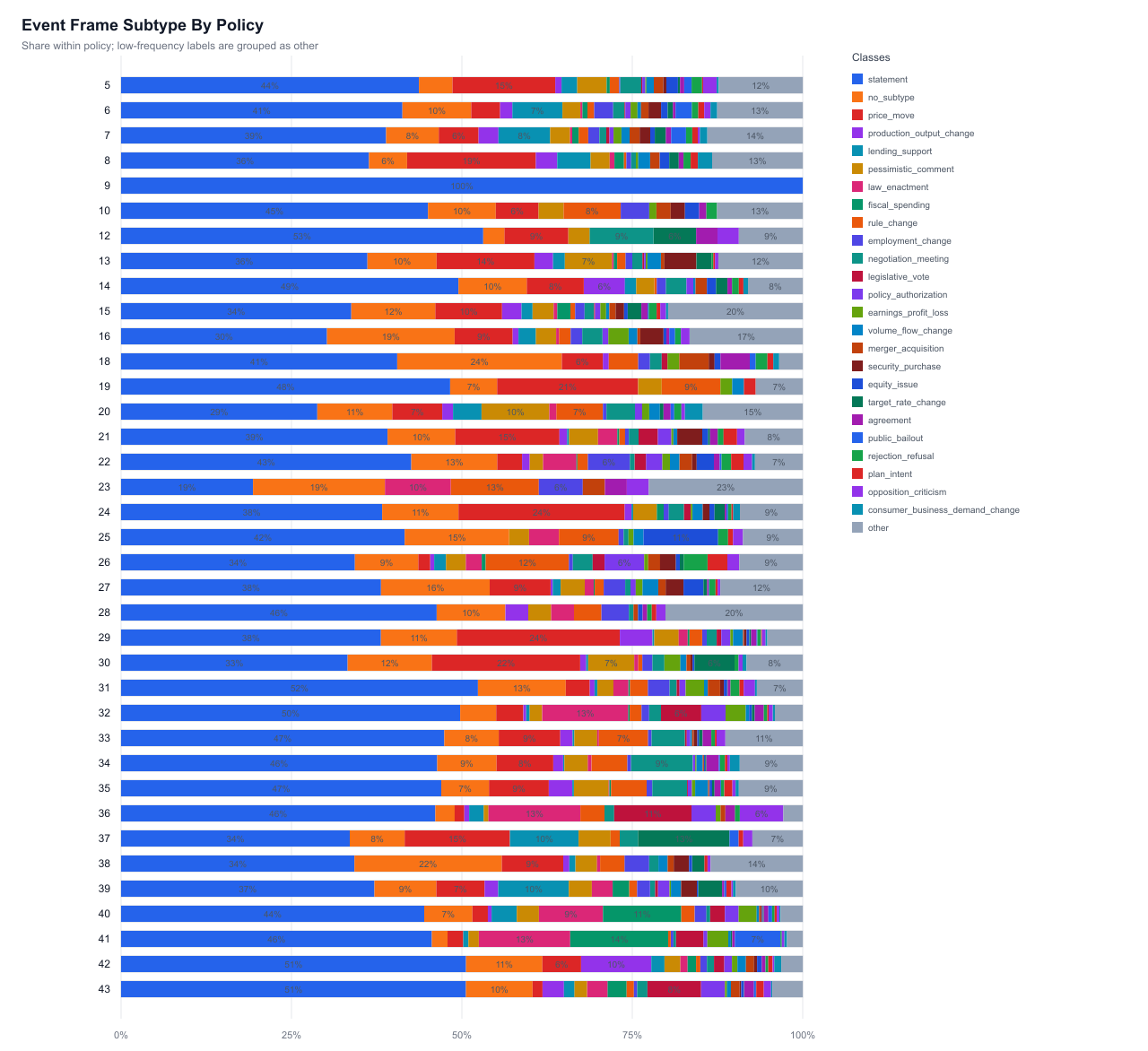}
\caption{Overall distribution of financial-action subtypes. Bars show the shares of the eight most frequent subtypes, with remaining subtypes pooled. Numeric codes correspond to the accompanying category key.}
    \label{fig:eventframe}
\end{figure}

\begin{table}[t]
\centering
\caption{Actor identity-resolution and action-row coverage. Entity-level percentages use 13,720 unique entity strings as the denominator, except Wikidata coverage, which uses 4,317 canonical labels. Row-level percentages use 31,460 action rows.}
\label{tab:actor-resolution-coverage}
\small
\setlength{\tabcolsep}{4pt}
\begin{tabularx}{\columnwidth}{@{}Xrr@{}}
\toprule
Measure & Count & Share \\
\midrule
\multicolumn{3}{@{}l}{\textit{Identity resolution}} \\
Unique nonblank entity strings
    & 13,720 & 100.0\% \\
Entity strings assigned to an organization
    & 7,810 & 56.9\% \\
Canonical organization labels
    & 4,317 & -- \\
Distinct Wikidata QIDs
    & 2,578 & 59.7\% \\
\midrule
\multicolumn{3}{@{}l}{\textit{Action and ActionFrame coverage}} \\
Action / ActionFrame rows
    & 31,460 / 31,460 & 100.0\% \\
Rows linked to an organization
    & 20,708 / 20,708 & 65.8\% \\
Rows linked to a Wikidata QID
    & 13,845 / 13,845 & 44.0\% \\
Rows without an organization
    & 10,752 / 10,752 & 34.2\% \\
\bottomrule
\end{tabularx}
\end{table}
\paragraph{Event-Frame Distributions} Figure~\ref{fig:policy_family} reveals financial-action family composition by policy while Figure~\ref{fig:eventframe} shows the overall distribution of financial-action subtypes

\begin{figure}[h]
    \centering
    \includegraphics[width=0.5\linewidth]{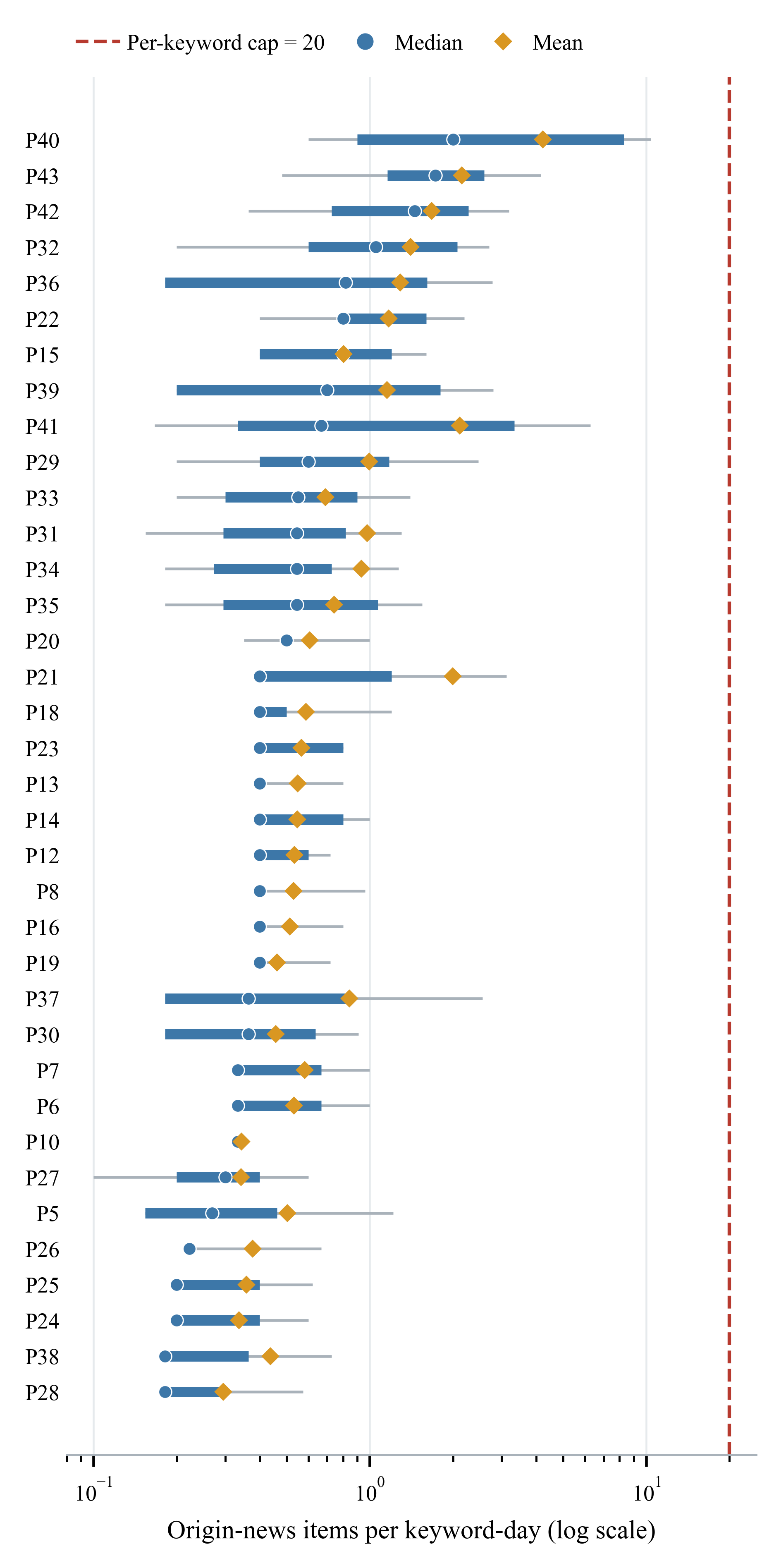}
\caption{Per-keyword daily origin-news distributions by policy. Dots denote medians, diamonds denote means, thick segments show interquartile ranges, and thin segments show the 10th--90th percentiles across observed policy-days. The dashed line marks the 20-item per-keyword reference.}
    \label{fig:news}
\end{figure}

\paragraph{News Information} Figure~\ref{fig:news} shows the per-keyword daily origin-news distributions by policy.

\begin{figure}[h]
    \centering
    \includegraphics[width=0.5\linewidth]{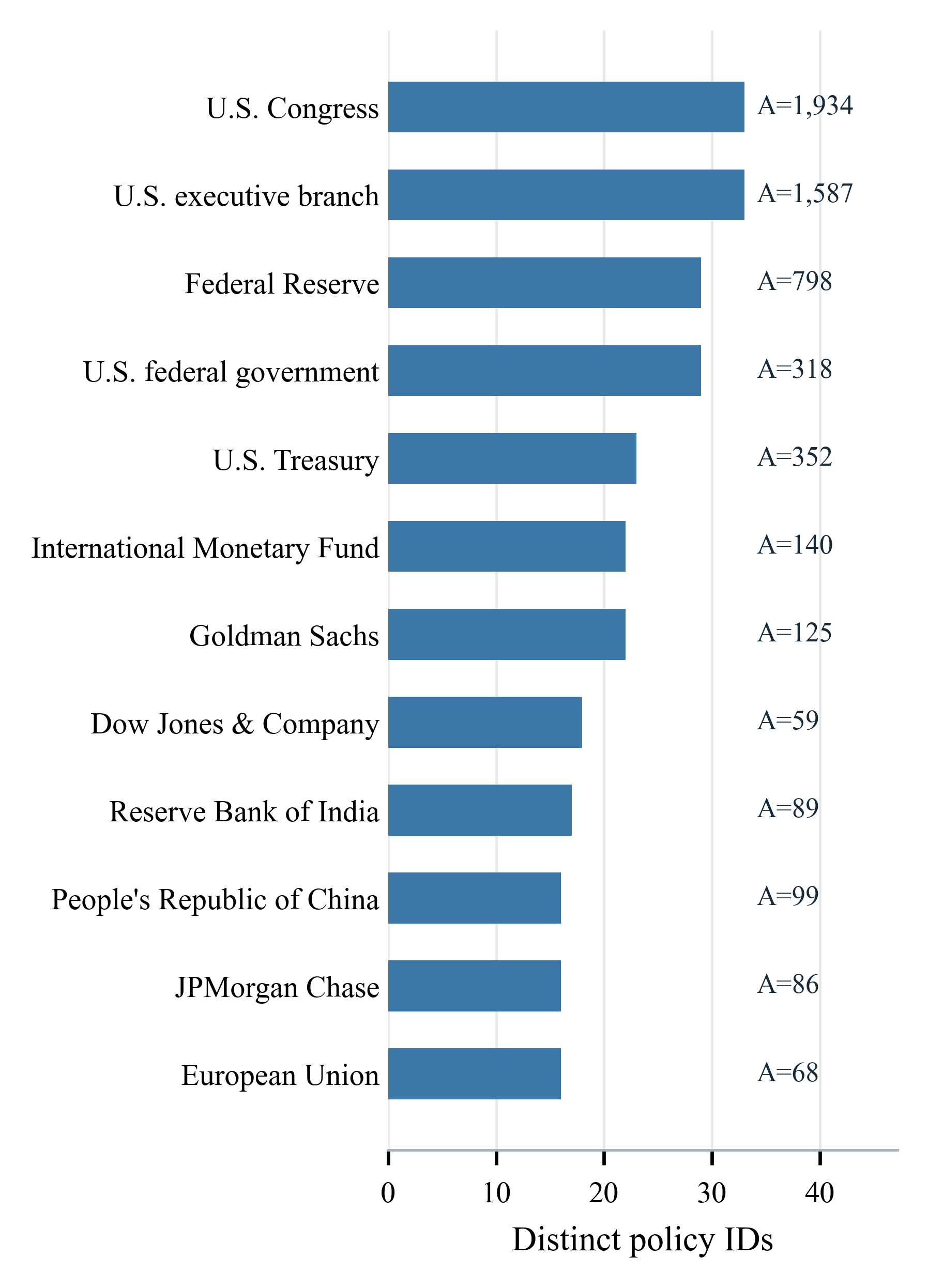}
\caption{Organisations with the widest policy coverage. Bar length shows the number of distinct policies, while \textit{A} reports the corresponding number of Action rows. Generic actors, countries, markets, and indices are excluded.}
    \label{fig:stakePop}
\end{figure}

\paragraph{Stakeholder Information} Figure~\ref{fig:stakePop} shows the organisations with the widest policy coverage.

\section{Detailed Case Study Data}~\label{app:case_study_detailed}

List of action\_ids of the US announcing and implementing the short-selling ban: [598, 679, 671, 674, 687, 696, 707, 712, 758, 776].

List of action\_ids of the UK announcing and implementing the short-selling ban: [625, 627, 686, 724, 672, 695, 702].

Some action\_ids showing positive returns for non-finance stocks: [677, 812, 817, 818, 820, 822, 1069, 1071, 1072].

Some action\_ids blaming short-sellers:[616, 648, 766, 797].

Some action\_ids opposed to the short-selling bans: [10744, 1154].

Action\_ids related to the MorganStanley retaliatory shorting: [1052, 1054, 1055].

Decimalization timeline: phase-in announcement (4391, 4392), delays (4399), NYSE completed (4468, 4469), Nasdaq completed (4470, 4400), US all completed (4401).

Action\_ids related to the suspicion of profit from fractional pricing: [4394, 4395, 4472, 4473].

Action\_ids related to decimal pricing benefits for investors: [4396, 4397, 4472, 4474].

\section{Additional Simulation Diagnostics}
\label{app:simulation_diagnostics}

This appendix reports compact simulation and feature-discovery diagnostics that support Section~\ref{sec:simulation-results}. We keep only plots that directly clarify replay difficulty, model calibration, or feature signal.

\section{Additional Comparison to Related Work}
\label{app:related-comparison}

Table~\ref{tab:related-comparison} summarizes how PAWS differs from existing simulators, financial NLP datasets, event-stream resources, and event-study databases. The comparison focuses on whether each line of work provides the ingredients needed for policy-driven multi-agent simulation: temporally grounded policy evidence, policy intent, stakeholder actions, entity relevance, and realized market context.

\begin{table*}[t]
\centering
\small
\setlength{\tabcolsep}{4pt}
\begin{tabular}{p{0.18\linewidth} p{0.23\linewidth} p{0.24\linewidth} p{0.27\linewidth}}
\hline
\textbf{Line of work} & \textbf{Representative examples} & \textbf{Primary focus} & \textbf{Limitation for policy-driven MAS} \\
\hline
Agent-based financial simulation
& ACF models, ABIDES, FinRL, TradeMaster~\citeapp{Lebaron2006,FarmerFoley2009,Byrd2019ABIDES,Liu_2021FinRL,Sun2023TradeMaster}
& Simulating trader behavior, order-book dynamics, market mechanisms, or reinforcement-learning trading policies.
& Provides simulation infrastructure, but typically does not curate historical policy episodes with policy intent, stakeholder responses, and realized market context. \\

LLM-agent simulation
& Generative Agents, SOTOPIA, QuantAgents, TradingAgents~\citeapp{Park2023GenerativeAgents,Zhou2023SOTOPIA,liuetal2025quantagents,Xiao2024TradingAgents}
& Modeling interactive agents that communicate, remember, plan, or collaborate in social and financial settings.
& Enables agent interaction, but still requires grounded evidence to evaluate whether simulated behavior matches real policy episodes. \\

Financial language resources
& FinGPT, FinQA, FinRED, Doc2EDAG, FINEED~\citeapp{Liuetal2023FINGPT,chen2022FinQA,Sharma2023FinRED,zheng-etal-2019-doc2edag,Huang2024FINEED}
& Financial text modeling, numerical reasoning, relation extraction, and document-level event extraction.
& Supports extraction and reasoning over financial text, but does not jointly align policy motivation, stakeholder actions, entity relevance, and market outcomes. \\

News event streams and timelines
& GDELT, ICEWS, EventRegistry, timeline summarization~\citeapp{WardEtAl2013GDELTICEWS,KwakAn2016GDELTEventRegistry,hu-etal-2024-moments,sunset}
& Broad event coverage, event clustering, temporal organization, and timeline generation from news streams.
& Captures event structure at scale, but is not specialized for finance-policy interventions or downstream market simulation. \\

Policy-shock and event-study resources
& Event studies, economic policy uncertainty, monetary-policy event-study databases~\citeapp{MacKinlay1997EventStudies,BakerBloomDavis2016EPU,AcostaEtAl2025USMPD}
& Estimating or organizing the market impact of discrete policy and economic events.
& Connects policy events to market responses, but is usually designed for treatment-effect estimation rather than stakeholder-level simulation traces. \\

\textbf{PAWS}
& This work
& Curated policy-centered episodes with policy intent, natural-language evidence, stakeholder actions, entity relevance, and financial-market context.
& Provides a temporally grounded substrate for studying policy-shock propagation in multi-agent financial simulation. \\
\hline
\end{tabular}
\caption{Comparison between PAWS and related work. PAWS is designed to support policy-driven multi-agent simulation by jointly preserving policy intent, stakeholder behavior, entity relevance, and realized market context.}
\label{tab:related-comparison}
\end{table*}

\section{Detailed Research Questions and Use Cases} \label{app:use-cases}

PAWS is motivated by research questions that require policy context, stakeholder actions, and market outcomes to be studied together. The dataset does not by itself solve causal identification in MAS, but it provides the temporally aligned evidence needed to define and evaluate such questions.

\paragraph{Agent-to-Agent Impact}
How reliably can agents affect one another in bottom-up simulations? This question is crucial for interpretable multiagent modelling, because a simulation should not only produce a final aggregate outcome but also explain which agents influenced which other agents, and when. PAWS supports this line of work by linking policy-relevant actions to dated stakeholders and organizations, enabling multi-turn analysis of agent-to-agent interactions. Future work can use these traces to study topological mechanisms for selective influence, where only certain agents are expected to affect specific downstream agents at particular timestamps.

\paragraph{Action-Outcome Alignment}
Disregarding the efficiency of agent-to-agent causality, can a multiagent system generate outcomes that are aligned with agentic actions? PAWS provides action types, actor identities, event dates, and surrounding policy context that can be used to test whether simulated outcomes are consistent with the actions agents take. This enables research on the parameters and design choices needed for aligned action-outcome generation, including agent memory, interaction topology, policy conditioning, and temporal update rules.

\paragraph{Expected and Unexpected Policy Outcomes}
How can multiagent AI systems robustly capture both expected and unexpected outcomes of policies? PAWS is designed around policies with documented intentions and observed market effects, allowing models to compare intended policy goals with subsequent stakeholder reactions and market context. This supports future work on counterfactual policy simulation, robustness under regime shifts, and the ability of agent populations to model unanticipated second-order effects.

\clearpage

\section{Additional Simulation Diagnostics}
\label{app:minimal_replay}
 We ran a full-process replay for the September 2008 short-selling ban over a larger selected agent set. This setting evaluates additional diffusion and opinion-dynamics baselines, including linear-threshold and independent-cascade models~\citeapp{Kempe2003Influence}, a voter model~\citeapp{HolleyLiggett1975Voter}, DeGroot consensus dynamics~\citeapp{DeGroot1974Consensus}, Bass diffusion~\citeapp{Bass1969NewProduct}, structured-classifier, and \texttt{GPT-4o} LLM-agent variants. Models are compared using trigger prediction, action-type accuracy, responder recall, timing error, action-distribution fit, and cascade-shape similarity.
 
This appendix reports compact simulation and feature-discovery diagnostics. We keep only plots that directly clarify replay difficulty, model calibration, or feature signal.

This simulation design is intentionally modular. Future work can replace the current decision rules with richer agent memory, role-conditioned prompts, selective agent-to-agent influence, counterfactual policy interventions, or explicit market-feedback modules. Thus, the methodology defines both the PAWS dataset construction procedure and the replay interface needed to test whether multi-agent systems can reproduce expected and unexpected responses to financial policy shocks.

\subsection{Minimal Multi-Policy Replay}
\label{app:minimal_replay}

Table~\ref{tab:minimal-replay-results} reports the \texttt{GPT-4o}-augmented minimal study over five policy events. The majority and transition baselines achieve high accuracy by predicting almost all rows as \emph{no action}, but their active-action F1 is near zero. The threshold and hybrid ABM variants recover some active days, while the structured classifier has the highest active recall among full-panel models. The bounded \texttt{GPT-4o} replay is evaluated only on a 50-row subset and is therefore not directly comparable by raw accuracy.

\begin{table*}[!t]
\centering
\small
\setlength{\tabcolsep}{6pt}
\renewcommand{\arraystretch}{1.08}
\begin{tabular}{l r r r r r}
\toprule
\textbf{Model} & \textbf{N} & \textbf{Acc.} & \textbf{Macro-F1} & \textbf{Active-F1} & \textbf{Active Recall} \\
\midrule
Majority & 8341 & .980 & .330 & .000 & .000 \\
Transition & 8341 & .979 & .330 & .012 & .006 \\
Threshold ABM & 8341 & .969 & .349 & .080 & .067 \\
Hybrid ABM & 8341 & .947 & .375 & .095 & .139 \\
Structured classifier & 8341 & .877 & .356 & .112 & .382 \\
\texttt{GPT-4o} replay & 50 & .240 & .205 & .718 & .560 \\
\bottomrule
\end{tabular}
\caption{Minimal multi-policy replay results from \texttt{minimal\_first\_study\_outputs\_llm}. The \texttt{GPT-4o} replay is bounded to a 50-row sample and should not be compared to full-panel models by raw accuracy.}
\label{tab:minimal-replay-results}
\end{table*}

\FloatBarrier

\noindent
\begin{minipage}{\linewidth}
    \centering
    \includegraphics[width=0.95\linewidth]{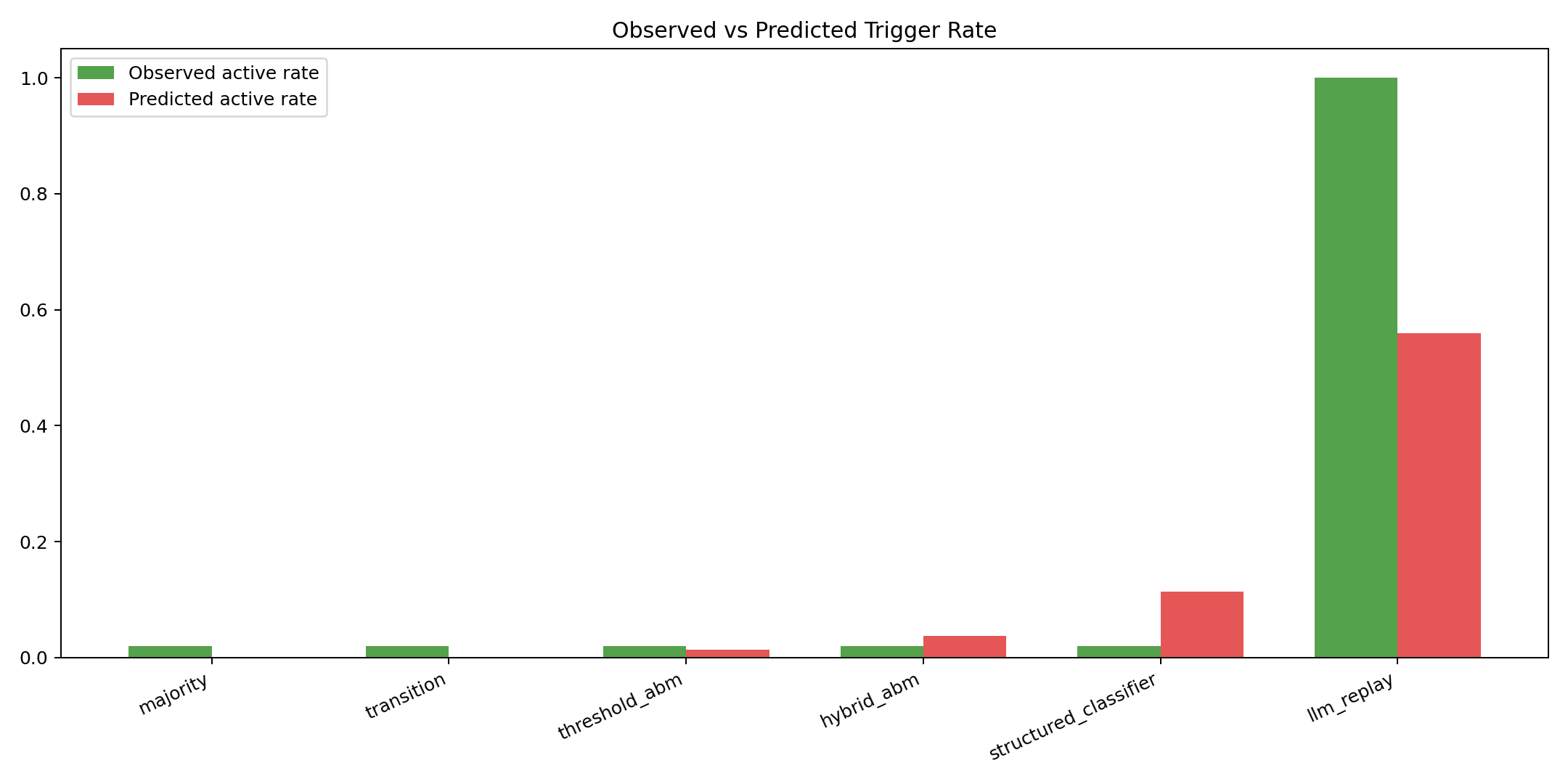}
    \captionof{figure}{Core metrics for the minimal multi-policy replay. This figure is moved from the main text because it supports the same imbalance argument as Table~\ref{tab:minimal-replay-results}.}
    \label{fig:minimal_core_metrics}
\end{minipage}

\vspace{1.2em}

% \noindent
% \begin{minipage}{\linewidth}
%     \centering
%     \includegraphics[width=0.95\linewidth]{imgs/simulation_policy5_core_metrics.png}
%     \captionof{figure}{Additional core metrics for the Policy 5 full-process replay. The main text uses the more compact action-recovery plot in Figure~\ref{fig:policy5_action_recovery}.}
%     \label{fig:policy5_core_metrics}
% \end{minipage}

% \FloatBarrier

The structured classifier provides the best active-F1 and macro-F1, while the \texttt{GPT-4o} LLM-agent variant obtains the highest active recall but substantially over-triggers. The Markov baseline produces the closest action-distribution fit, suggesting that simple temporal persistence can match aggregate action rates even when it misses many specific responders.
Table~\ref{tab:minimal-replay-results} reports the \texttt{GPT-4o}-augmented minimal study over policy five. The majority and transition baselines achieve high accuracy by predicting almost all rows as \emph{no action}, but their active-action F1 is near zero. The threshold and hybrid ABM variants recover some active days, while the structured classifier has the highest active recall among full-panel models. The bounded \texttt{GPT-4o} replay is evaluated only on a 50-row subset and is therefore not directly comparable by raw accuracy. The post-calibration Policy 5 split in the same output directory contains 560 panel rows and confirms this interpretation. No-action diffusion baselines again reach high accuracy (.939) but zero active-F1. The \texttt{GPT-4o} LLM-agent variant reaches .824 active recall but only .122 active-F1, while the structured classifier is more conservative, with .069 active-F1 and .176 active recall. These results suggest that calibration remains the central challenge for action-triggering models.
\paragraph{Feature-discovery diagnostics}
We also run a feature-discovery study over the LLM-augmented panel using balanced logistic regression, L1-regularized logistic regression, and random forests~\citep{Tibshirani1996Lasso,Breiman2001RandomForests,Pedregosa2011ScikitLearn}. Rather than introducing another simulator, this study diagnoses which PAWS fields are predictive for downstream simulation tasks. 
% Figure~\ref{fig:feature_discovery_best_task_summar} shows that trigger prediction is the most learnable task, with balanced accuracy of .696 and macro-F1 of .498. Action-type prediction is also learnable when using agent identity features, while sentiment remains close to chance, consistent with the validation results in Section~\ref{sec:dataset_validation}.

% \begin{figure}[t]
%     \centering
%     \includegraphics[width=0.8\linewidth]{imgs/feature_discovery_best_task_summary.pdf}
%     \caption{Best feature-discovery model for each downstream task. Trigger prediction benefits from temporal and network features, action-type prediction is strongest with agent identity, and sentiment remains weak.}
%     \label{fig:feature_discovery_best_task_summary}
% \end{figure}

The top-ranked discovered features further clarify the dataset signal. For trigger prediction, the highest-ranked features are \texttt{org\_id}, previous label, weighted degree, neighbor context score, and policy identity. For action-type prediction, the leading features are \texttt{org\_id}, article count, recent policy activity, recent policy context activity, and previous label. For sentiment, the leading features are \texttt{org\_id}, weighted degree, relevance score, policy-level intervention count, and policy identity. These diagnostics support placing feature discovery in the simulation-results section: it evaluates the replay interface by identifying which dataset fields models actually use.

\subsection{Feature-Discovery Diagnostics}
\label{app:feature_diagnostics}

\begin{figure*}[t]
    \centering
    \begin{subfigure}{0.49\textwidth}
        \centering
        \includegraphics[width=\linewidth]{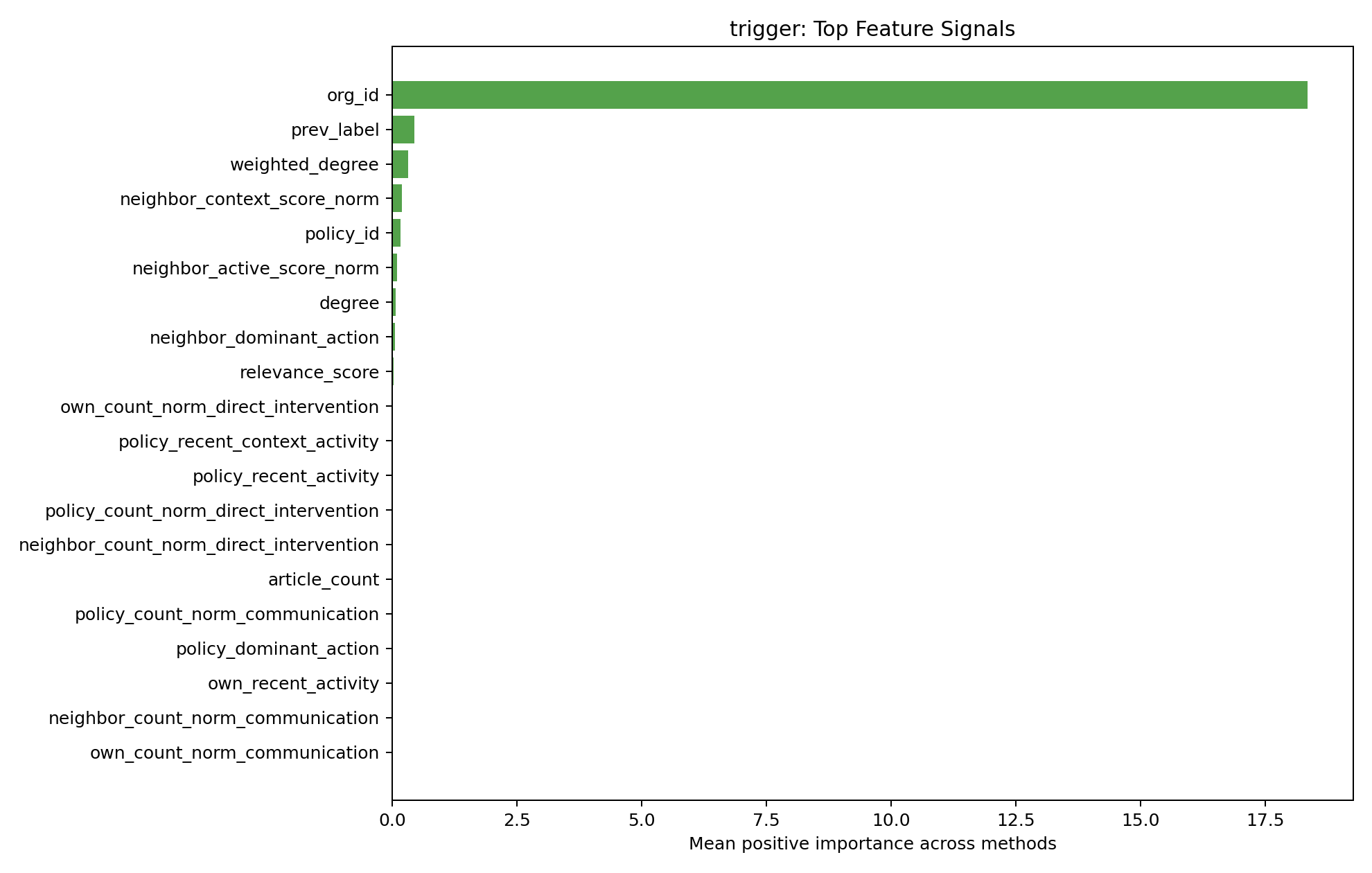}
        \caption{Top feature signals}
    \end{subfigure}
    \hfill
    \begin{subfigure}{0.49\textwidth}
        \centering
        \includegraphics[width=\linewidth]{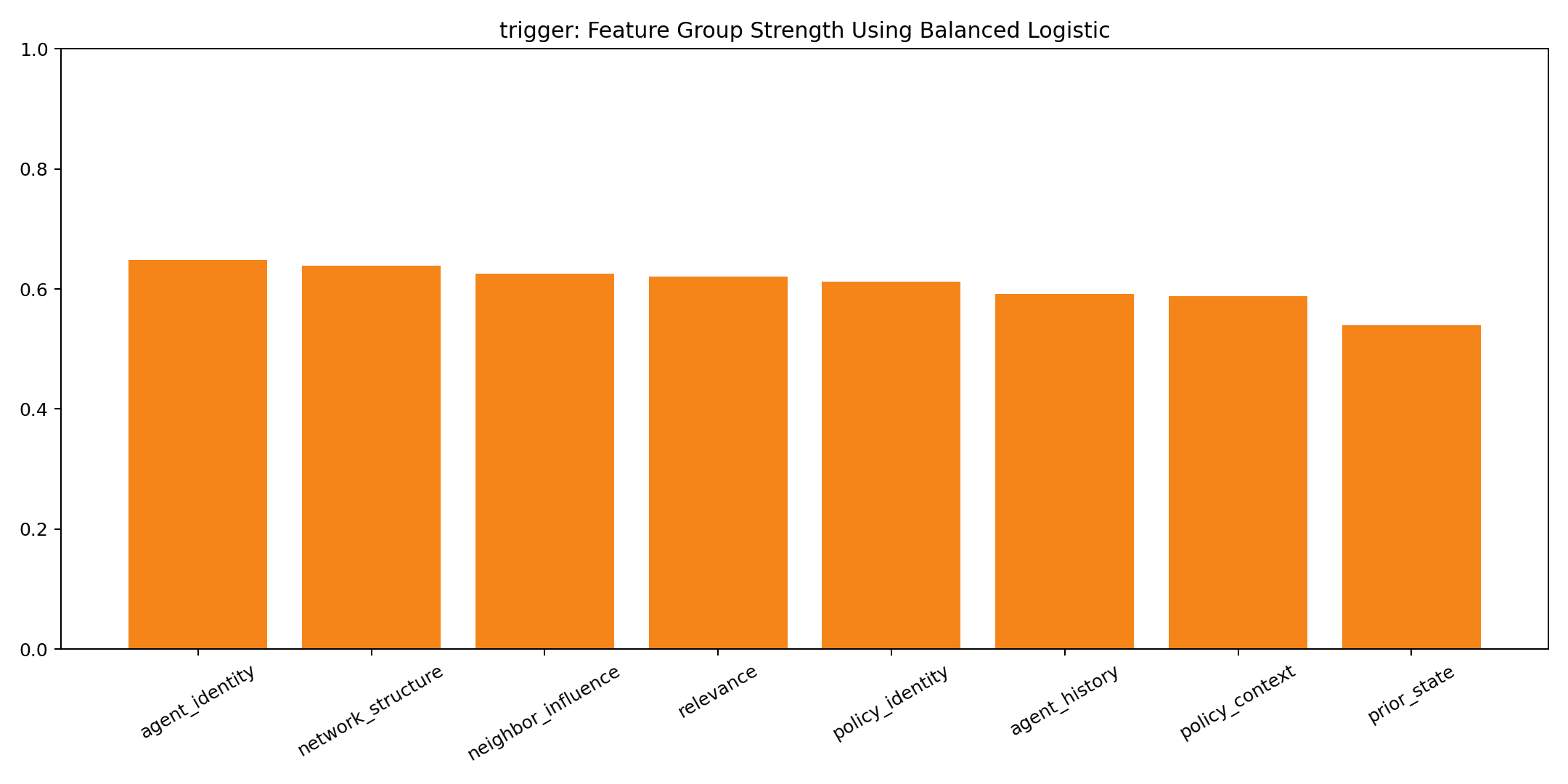}
        \caption{Feature-group strength}
    \end{subfigure}
    \caption{Feature-discovery diagnostics for sentiment prediction. The weak and diffuse signal is consistent with the validation finding that policy-event sentiment is often target-dependent and unstable.}
    \label{fig:feature_sentiment_diagnostics}
\end{figure*}

Figures~\ref{fig:feature_sentiment_diagnostics}--\ref{fig:feature_trigger_diagnostics} provide compact feature-discovery diagnostics for the replay tasks. These plots identify which input groups carry the strongest signal for action triggering, action-type prediction, and sentiment prediction.

\begin{figure*}[t]
    \centering
    \begin{subfigure}{0.49\textwidth}
        \centering
        \includegraphics[width=\linewidth]{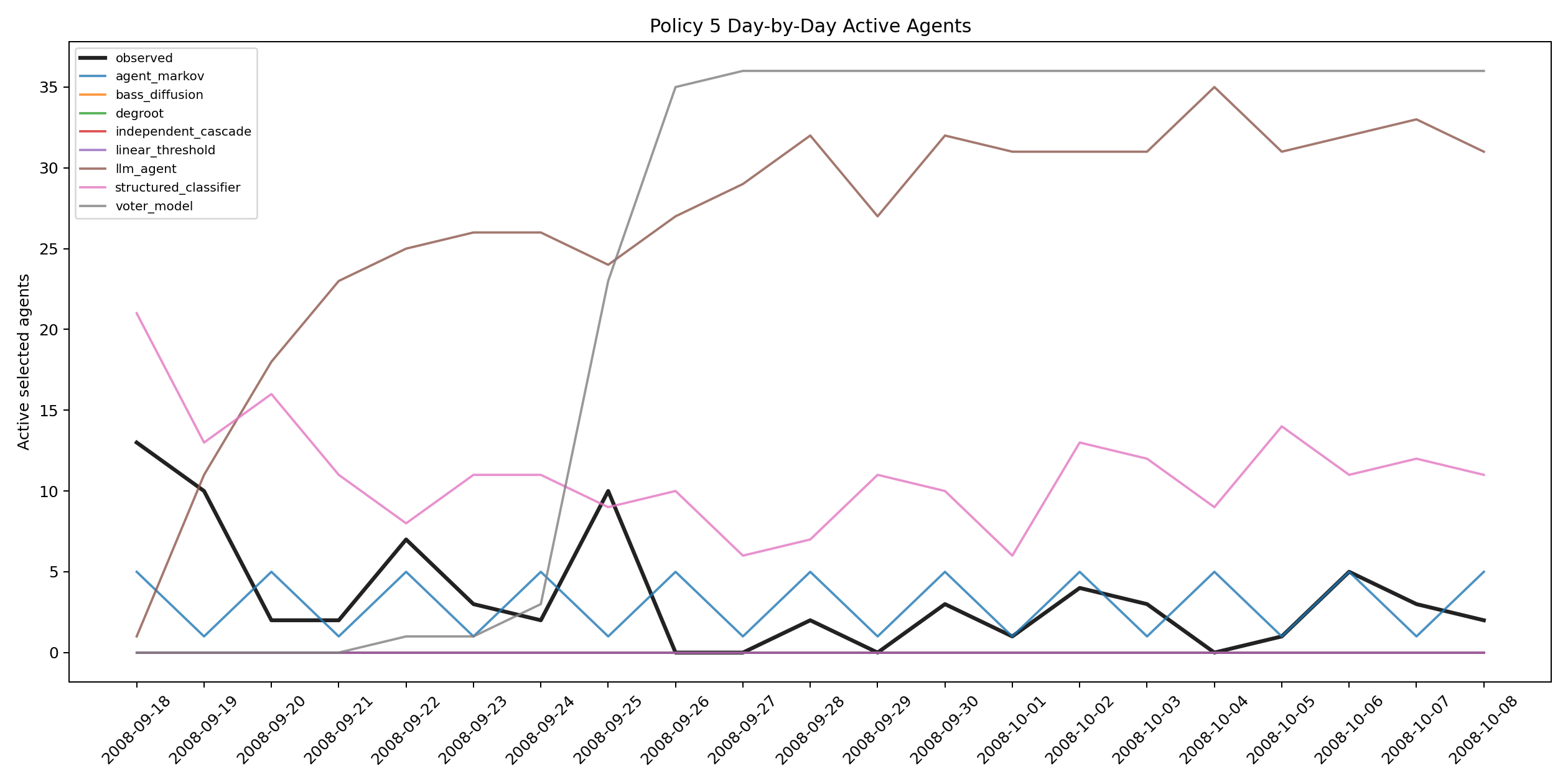}
        \caption{Top feature signals}
    \end{subfigure}
    \hfill
    \begin{subfigure}{0.49\textwidth}
        \centering
        \includegraphics[width=\linewidth]{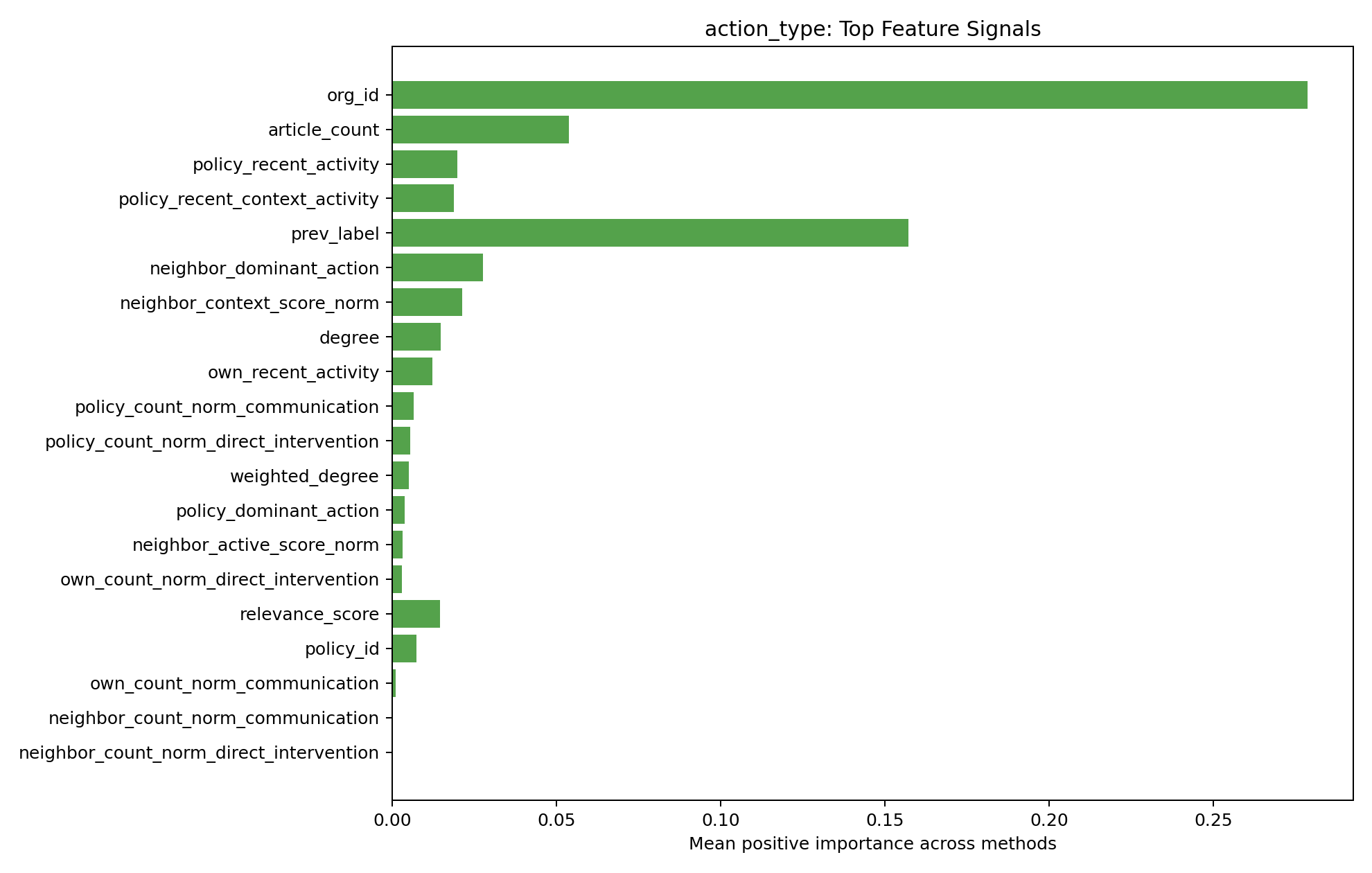}
        \caption{Feature-group strength}
    \end{subfigure}
    \caption{Feature-discovery diagnostics for action triggering. Identity, prior state, network structure, and neighbor context provide the strongest trigger signal.}
    \label{fig:feature_trigger_diagnostics}
\end{figure*}

\begin{figure*}[t]
    \centering
    \begin{subfigure}{0.49\textwidth}
        \centering
        \includegraphics[width=\linewidth]{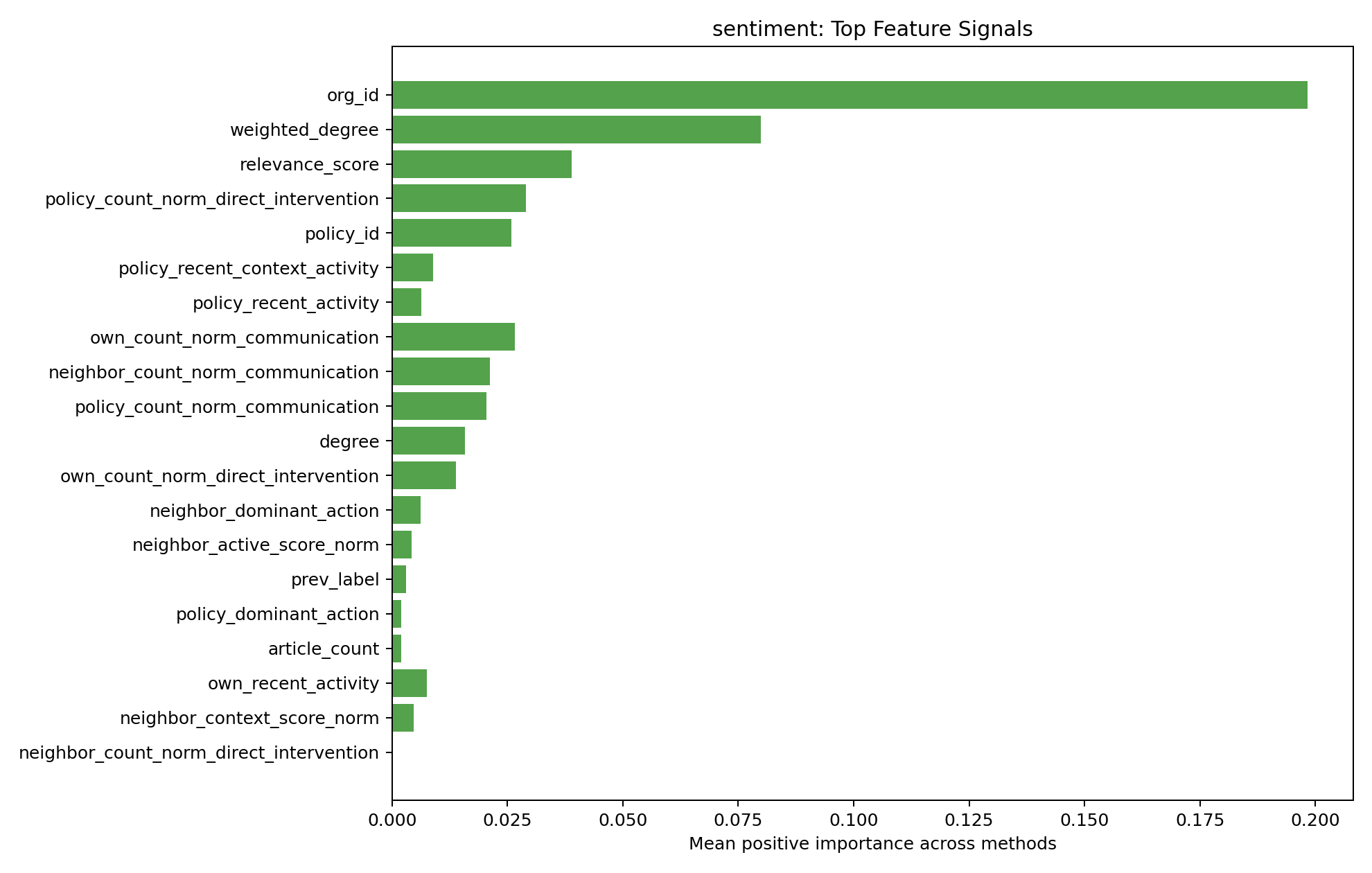}
        \caption{Top feature signals}
    \end{subfigure}
    \hfill
    \begin{subfigure}{0.49\textwidth}
        \centering
        \includegraphics[width=\linewidth]{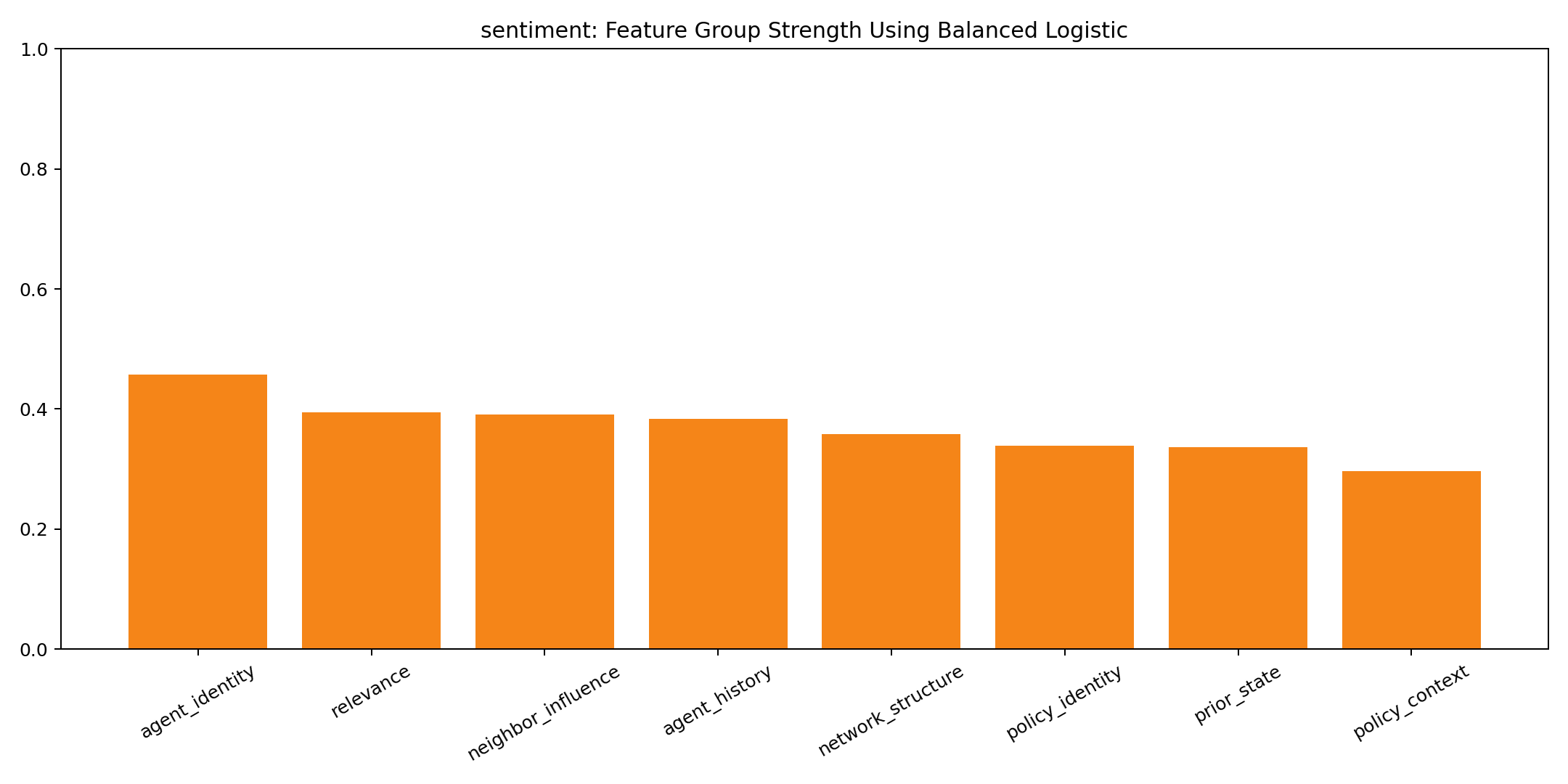}
        \caption{Feature-group strength}
    \end{subfigure}
    \caption{Feature-discovery diagnostics for action-type prediction. Agent identity and article-frequency features dominate, indicating that organizations have stable action-mode tendencies in the current panel.}
    \label{fig:feature_action_type_diagnostics}
\end{figure*}

\begin{figure}[!htbp]
    \centering
    \begin{minipage}{0.98\linewidth}
        \centering
        \includegraphics[width=\linewidth]{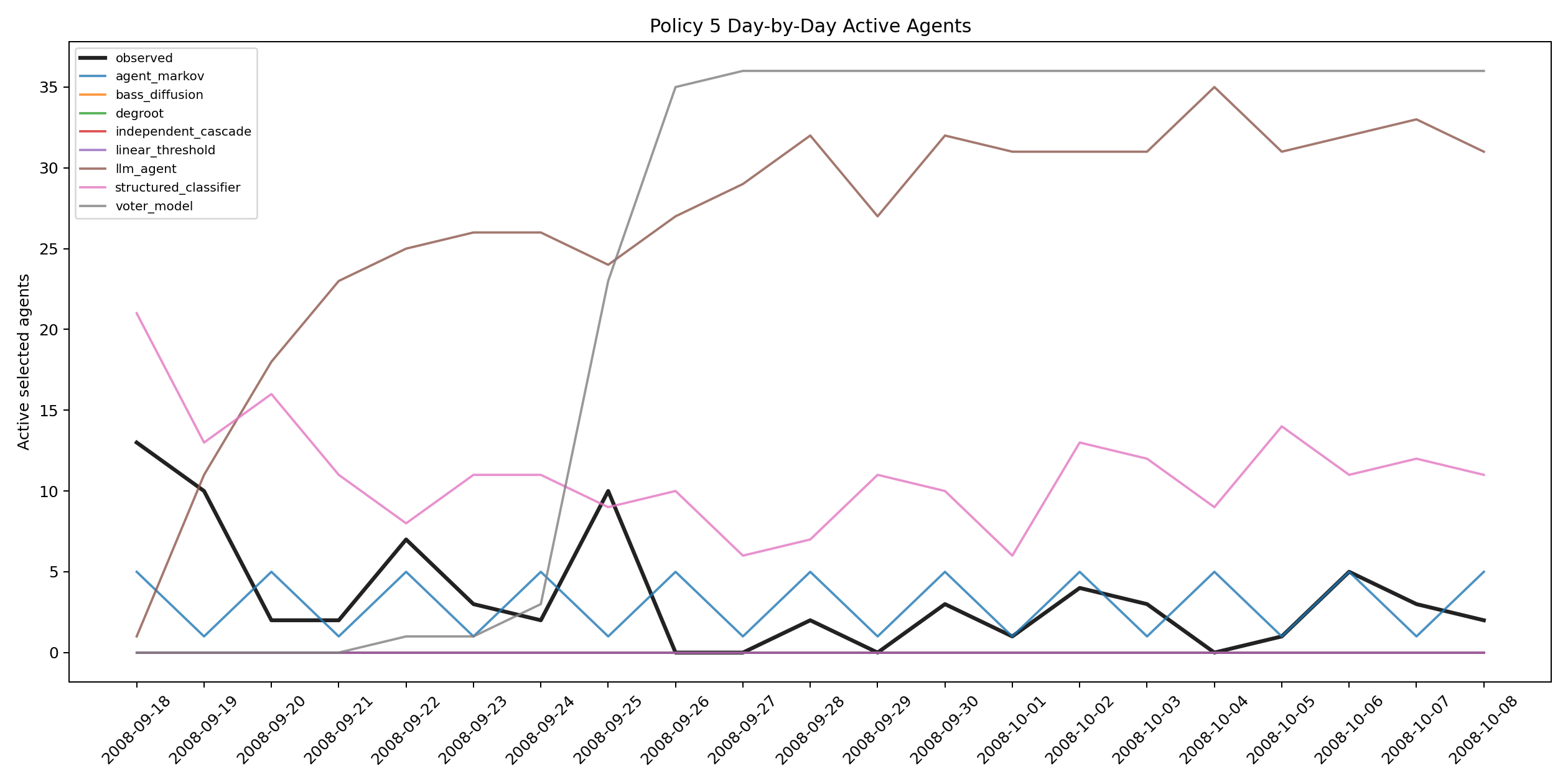}
        \caption*{(a) Top feature signals}
    \end{minipage}

    \vspace{0.6em}

    \begin{minipage}{0.9\linewidth}
        \centering
        \includegraphics[width=\linewidth]{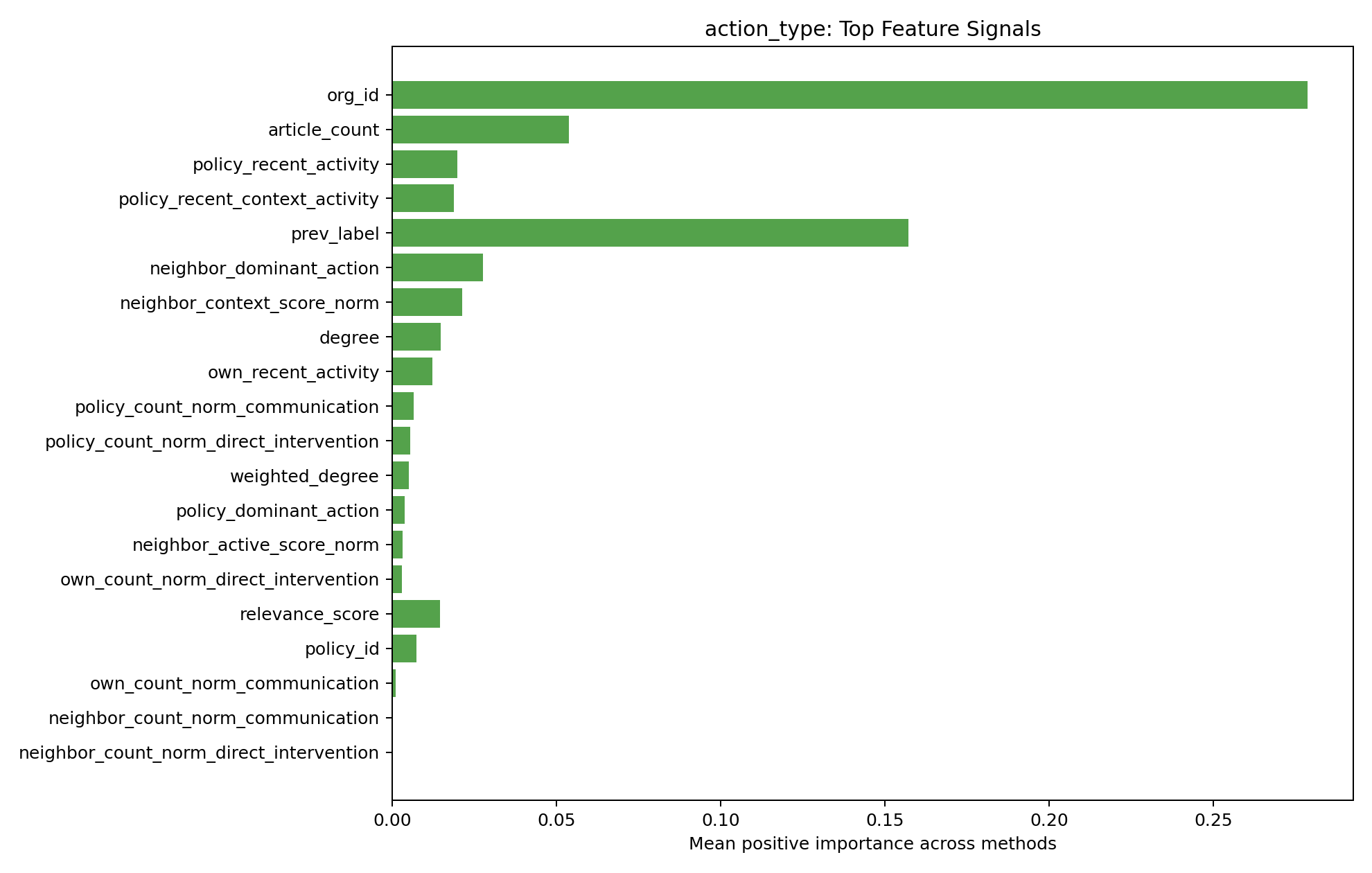}
        \caption*{(b) Feature-group strength}
    \end{minipage}

    \caption{Feature-discovery diagnostics for action triggering. Identity, prior state, network structure, and neighbor context provide the strongest trigger signal.}
    \label{fig:feature_trigger_diagnostics}
\end{figure}

\FloatBarrier
\clearpage

\section{Additional Comparison to Related Work}
\label{app:related-comparison}

Table~\ref{tab:related-comparison} summarizes how PAWS differs from existing simulators, financial NLP datasets, event-stream resources, and event-study databases. The comparison focuses on whether each line of work provides the ingredients needed for policy-driven multi-agent simulation: temporally grounded policy evidence, policy intent, stakeholder actions, entity relevance, and realized market context.

\begin{table*}[!t]
\centering
\small
\setlength{\tabcolsep}{4pt}
\renewcommand{\arraystretch}{1.08}
\begin{tabular}{p{0.18\linewidth} p{0.23\linewidth} p{0.24\linewidth} p{0.27\linewidth}}
\hline
\textbf{Line of work} & \textbf{Representative examples} & \textbf{Primary focus} & \textbf{Limitation for policy-driven MAS} \\
\hline
Agent-based financial simulation
& ACF models, ABIDES, FinRL, TradeMaster~\citeapp{Lebaron2006,FarmerFoley2009,Byrd2019ABIDES,Liu_2021FinRL,Sun2023TradeMaster}
& Simulating trader behavior, order-book dynamics, market mechanisms, or reinforcement-learning trading policies.
& Provides simulation infrastructure, but typically does not curate historical policy episodes with policy intent, stakeholder responses, and realized market context. \\

LLM-agent simulation
& Generative Agents, SOTOPIA, QuantAgents, TradingAgents~\citeapp{Park2023GenerativeAgents,Zhou2023SOTOPIA,liuetal2025quantagents,Xiao2024TradingAgents}
& Modeling interactive agents that communicate, remember, plan, or collaborate in social and financial settings.
& Enables agent interaction, but still requires grounded evidence to evaluate whether simulated behavior matches real policy episodes. \\

Financial language resources
& FinGPT, FinQA, FinRED, Doc2EDAG, FINEED~\citeapp{Liuetal2023FINGPT,chen2022FinQA,Sharma2023FinRED,zheng-etal-2019-doc2edag,Huang2024FINEED}
& Financial text modeling, numerical reasoning, relation extraction, and document-level event extraction.
& Supports extraction and reasoning over financial text, but does not jointly align policy motivation, stakeholder actions, entity relevance, and market outcomes. \\

News event streams and timelines
& GDELT, ICEWS, EventRegistry, timeline summarization~\citeapp{WardEtAl2013GDELTICEWS,KwakAn2016GDELTEventRegistry,hu-etal-2024-moments,sunset}
& Broad event coverage, event clustering, temporal organization, and timeline generation from news streams.
& Captures event structure at scale, but is not specialized for finance-policy interventions or downstream market simulation. \\

Policy-shock and event-study resources
& Event studies, economic policy uncertainty, monetary-policy event-study databases~\citeapp{MacKinlay1997EventStudies,BakerBloomDavis2016EPU,AcostaEtAl2025USMPD}
& Estimating or organizing the market impact of discrete policy and economic events.
& Connects policy events to market responses, but is usually designed for treatment-effect estimation rather than stakeholder-level simulation traces. \\

\textbf{PAWS}
& This work
& Curated policy-centered episodes with policy intent, natural-language evidence, stakeholder actions, entity relevance, and financial-market context.
& Provides a temporally grounded substrate for studying policy-shock propagation in multi-agent financial simulation. \\
\hline
\end{tabular}
\caption{Comparison between PAWS and related work. PAWS is designed to support policy-driven multi-agent simulation by jointly preserving policy intent, stakeholder behavior, entity relevance, and realized market context.}
\label{tab:related-comparison}
\end{table*}

\FloatBarrier
\clearpage

\section{Detailed Research Questions and Use Cases}
\label{app:use-cases}

PAWS is motivated by research questions that require policy context, stakeholder actions, and market outcomes to be studied together. The dataset does not by itself solve causal identification in MAS, but it provides the temporally aligned evidence needed to define and evaluate such questions.

\paragraph{Agent-to-Agent Impact}
How reliably can agents affect one another in bottom-up simulations? This question is crucial for interpretable multiagent modelling, because a simulation should not only produce a final aggregate outcome but also explain which agents influenced which other agents, and when. PAWS supports this line of work by linking policy-relevant actions to dated stakeholders and organizations, enabling multi-turn analysis of agent-to-agent interactions. Future work can use these traces to study topological mechanisms for selective influence, where only certain agents are expected to affect specific downstream agents at particular timestamps.

\paragraph{Action-Outcome Alignment}
Disregarding the efficiency of agent-to-agent causality, can a multiagent system generate outcomes that are aligned with agentic actions? PAWS provides action types, actor identities, event dates, and surrounding policy context that can be used to test whether simulated outcomes are consistent with the actions agents take. This enables research on the parameters and design choices needed for aligned action-outcome generation, including agent memory, interaction topology, policy conditioning, and temporal update rules.

\paragraph{Expected and Unexpected Policy Outcomes}
How can multiagent AI systems robustly capture both expected and unexpected outcomes of policies? PAWS is designed around policies with documented intentions and observed market effects, allowing models to compare intended policy goals with subsequent stakeholder reactions and market context. This supports future work on counterfactual policy simulation, robustness under regime shifts, and the ability of agent populations to model unanticipated second-order effects.
\clearpage

\section{Complete EventFrame Category Inventory}\label{appn:frameclass}
% Generated by helper_func/generate_actionframe_paper_latex.py
\providecommand{\NA}{\textsc{n/a}}
\noindent\textit{Applicability convention.} The stored value \texttt{not\_applicable}, shown hereafter as \NA, denotes structural ineligibility rather than missingness or uncertainty. It is available to every independently gated crosswalk (CAMEO, Federal Reserve tools, ACE/ERE, monetary policy, iMaPP, and IMF crisis response) and, where appropriate, to status and direction. CAMEO and ACE child fields inherit \NA\ from their parent. To avoid ten repetitive rows, this shared value is declared once here and omitted from the inventory below.

\begingroup
\scriptsize
\setlength{\tabcolsep}{2.5pt}
\renewcommand{\arraystretch}{1.12}
\begin{longtable}{@{}p{0.13\textwidth}p{0.22\textwidth}p{0.20\textwidth}p{0.29\textwidth}p{0.12\textwidth}@{}}
\caption{Substantive category inventory for the EventFrame representation. The shared structural \NA\ value is defined above rather than repeated under every eligible field. Proposed layers are not yet adjudicated; external crosswalks are independently gated.}
\label{tab:eventframe-category-inventory}\\
\toprule
Layer & Category & Parent or constraint & Meaning & Provenance \\
\midrule
\endfirsthead
\multicolumn{5}{c}{\tablename\ \thetable{} --- continued} \\
\toprule
Layer & Category & Parent or constraint & Meaning & Provenance \\
\midrule
\endhead
\midrule
\multicolumn{5}{r}{Continued on next page} \\
\endfoot
\bottomrule
\endlastfoot
\texttt{event\_\allowbreak{}route (proposed)} & \texttt{agent\_\allowbreak{}action} & \textemdash & An intentional choice attributable to an actor and potentially available in the MAS action space. & Project synthesis from BDI/planning \\
\texttt{event\_\allowbreak{}route (proposed)} & \texttt{environment\_\allowbreak{}state\_\allowbreak{}or\_\allowbreak{}outcome} & \textemdash & An observed transition or outcome that normally belongs to the environment rather than an actor's selectable action set. & Project synthesis from BDI/planning \\
\addlinespace[2pt]
\texttt{underlying\_\allowbreak{}policy\_\allowbreak{}function (proposed)} & \texttt{information\_\allowbreak{}signal\_\allowbreak{}or\_\allowbreak{}disclosure} & \textemdash & Provide, disclose, forecast, warn, endorse, criticize, or otherwise change the information available to other agents. & Project synthesis: frames, policy tools, MAS \\
\texttt{underlying\_\allowbreak{}policy\_\allowbreak{}function (proposed)} & \texttt{influence\_\allowbreak{}request\_\allowbreak{}or\_\allowbreak{}pressure} & \textemdash & Try to induce another actor to act or refrain from acting through an appeal, request, demand, rejection, threat, or protest. & Project synthesis: frames, policy tools, MAS \\
\texttt{underlying\_\allowbreak{}policy\_\allowbreak{}function (proposed)} & \texttt{coordination\_\allowbreak{}consultation\_\allowbreak{}or\_\allowbreak{}commitment} & \textemdash & Consult, negotiate, agree, or express a commitment to cooperate. When a concrete embedded policy action exists, that policy function should supersede this fallback. & Project synthesis: frames, policy tools, MAS \\
\texttt{underlying\_\allowbreak{}policy\_\allowbreak{}function (proposed)} & \texttt{monetary\_\allowbreak{}stance} & \textemdash & Set or signal the current/future monetary-policy stance. & Project synthesis: frames, policy tools, MAS \\
\end{longtable}
\endgroup

\bibliographystyleapp{aaai2027}
\bibliographyapp{custom} % This imports the appendix.tex file

\end{document}